\documentclass[12pt]{article}

\usepackage{newtxtext,newtxmath}

\usepackage{graphicx}
\usepackage{multirow}

\usepackage{xcolor}
\usepackage{makecell}
\usepackage{pifont}
\usepackage[letterpaper,margin=1in]{geometry}

\renewenvironment{abstract}
	{\quotation}
	{\endquotation}

\date{}

\makeatletter
\renewcommand{\fnum@figure}{\textbf{Figure \thefigure}}
\renewcommand{\fnum@table}{\textbf{Table \thetable}}
\makeatother

\usepackage{scicite}

\usepackage{url}

\def\scititle{Tianmu-TC: Physics-constraints Generative Artificial Intelligence for Global Tropical Cyclone Forecasting}
\title{\bfseries \boldmath \scititle}

\author{
	Shiqi~Zhang$^{1\dagger}$,
	Pan~Mu$^{1\dagger}$,
	Cheng~Huang$^{1}$,
	Hanting~Yan$^{1}$,
	Yuchao~Zhu$^{1}$,\\
	Jinglin~Zhang$^{2}$,
	Shengyong~Chen$^{3\ast}$,
	Shoujuan~Shu$^{4\ast}$,
	Cong~Bai$^{1\ast}$\and
	\small$^{1}$College of Computer Science and Technology, Zhejiang University of Technology, Hangzhou \& 310023, China.\and
	\small$^{2}$School of Control Science and Engineering, Shandong University, Jinan \& 250061, China.\and
	\small$^{3}$School of Computer Science and Engineering, Tianjin University of Technology, Tianjin \& 300384, China.\and
	\small$^{4}$School of Earth Sciences, Zhejiang University, Hangzhou \& 310058, China.\and
	\small$^\ast$Corresponding author. Email: csy@tjut.edu.cn, sjshu@zju.edu.cn, congbai@zjut.edu.cn\and
	\small$^\dagger$These authors contributed equally to this work.
}

\begin{document} 

\maketitle

\begin{abstract} \bfseries \boldmath
Tropical cyclones (TCs) pose severe risks from strong winds and heavy rainfall. However, forecasting their track and intensity remains challenging due to chaotic atmosphere and the rapid amplification of initial condition errors, leading to growing forecast uncertainty. While numerical weather prediction (NWP) and deep learning models have made progress, they remain computationally demanding and often fail under complex meteorological scenarios. Here, we present Tianmu-TC, a physics-constraints generative framework for global TC forecasting. Trained on Western North Pacific data, Tianmu-TC leverages physics-constraints to generate controllable outputs with reduced uncertainty thus improving forecast reliability. Experiments show Tianmu-TC outperforms deterministic and ensemble meteorological artificial intelligence models and authoritative NWP systems such as ECMWF in global ocean basins, with significantly lower computational cost. We further show Tianmu-TC performs well in challenging scenarios such as data sparsity, anomaly tracks, rapid intensification and weakening. These findings suggest physics-constraints generative AI offers a promising approach for reliable, efficient global TC forecasting.

\end{abstract}

\noindent
\section*{Introduction}\label{sec1}
Tropical cyclones (TCs) are complex weather systems that often bring strong winds and heavy rainfall, leading to natural disasters that cause significant damage to infrastructure and pose serious threats to human life and property~\cite{knutson2010tropical}. Therefore, accurate forecasting is crucial for effective prevention of disasters caused by TCs. The track of a TC tells us where it is currently located, while its intensity represents its strength and the potential severity of the damage it may cause. Forecasting these two attributes are crucial for determining the level of preparedness required. Specifically, the TC track is a combination of locations, including longitude and latitude of the TC center. TC intensity can represented by different attributes, with pressure and wind speed being the most representative. Pressure refers to the lowest atmospheric pressure (hPa) at the TC center, and wind speed to the maximum sustained wind speed (m/s) over two minutes near the TC center. These attributes are closely interconnected in physics~\cite{callaghan1998relationship,knaff2007reexamination}.

However, forecasting tropical cyclones inherently involves uncertainty, making accurate predictions a persistent challenge. The atmosphere is a chaotic system, meaning that even small errors in the initial conditions can amplify over time—a phenomenon widely known as the butterfly effect—ultimately degrading forecast accuracy and increasing uncertainty. This uncertainty becomes especially pronounced as the forecast horizon extends, with both track and intensity predictions becoming less reliable. For instance, the predicted location of the TC center may deviate by several hundred kilometers, while intensity forecasts often fail to capture abrupt changes such as \textit{rapid intensification} or \textit{rapid weakening}. In practice, this leads to wide forecast spreads and rapidly growing uncertainty.

To address these challenges, authoritative Numerical Weather Prediction (NWP) models~\cite{bauer2015quiet,kimura2002numerical,nhcverify} adopt ensemble forecasting strategies, which simulate multiple scenarios by perturbing the initial conditions~\cite{leutbecher2008ensemble}. This approach has been widely adopted by official agencies such as the European Centre for Medium-Range Weather Forecasts (ECMWF)~\cite{ecmwf2011ifs}, the National Hurricane Center (OFCL)~\cite{nhcverify}, and the China Central Meteorological Observatory (CMO)~\cite{CMO}. While ensemble methods help mitigate uncertainty, they require substantial computational resources to simulate global atmospheric processes~\cite{sanders1975barotropic}, resulting in prohibitively long training and inference times. This makes them less suitable for real-time regional applications where rapid response is essential. 
As the number of uncertainties being considered grows, the required computational resources increase significantly. Thus, it is imperative to explore more efficient approaches for TC forecasting.

Recently, the advancement of generative models has provided a novel perspective for addressing the challenges of uncertainty~\cite{price2025probabilistic,gao2023prediff,liu2026uncertainty,chen2025generating,mardani2025residual} with substantially lower computational cost. By introducing controlled noise or perturbations, they technically circumvent the impact of initial errors and their iterative amplification, enabling the generation of future TC tracks and intensities. Furthermore, their ability to generate a large number of diverse members, similar to ensemble forecasting, makes them highly suitable for predicting future tropical cyclone tracks and intensities.

\begin{figure}[!htb]    
	\centering 
	\includegraphics[width=1.00\textwidth]{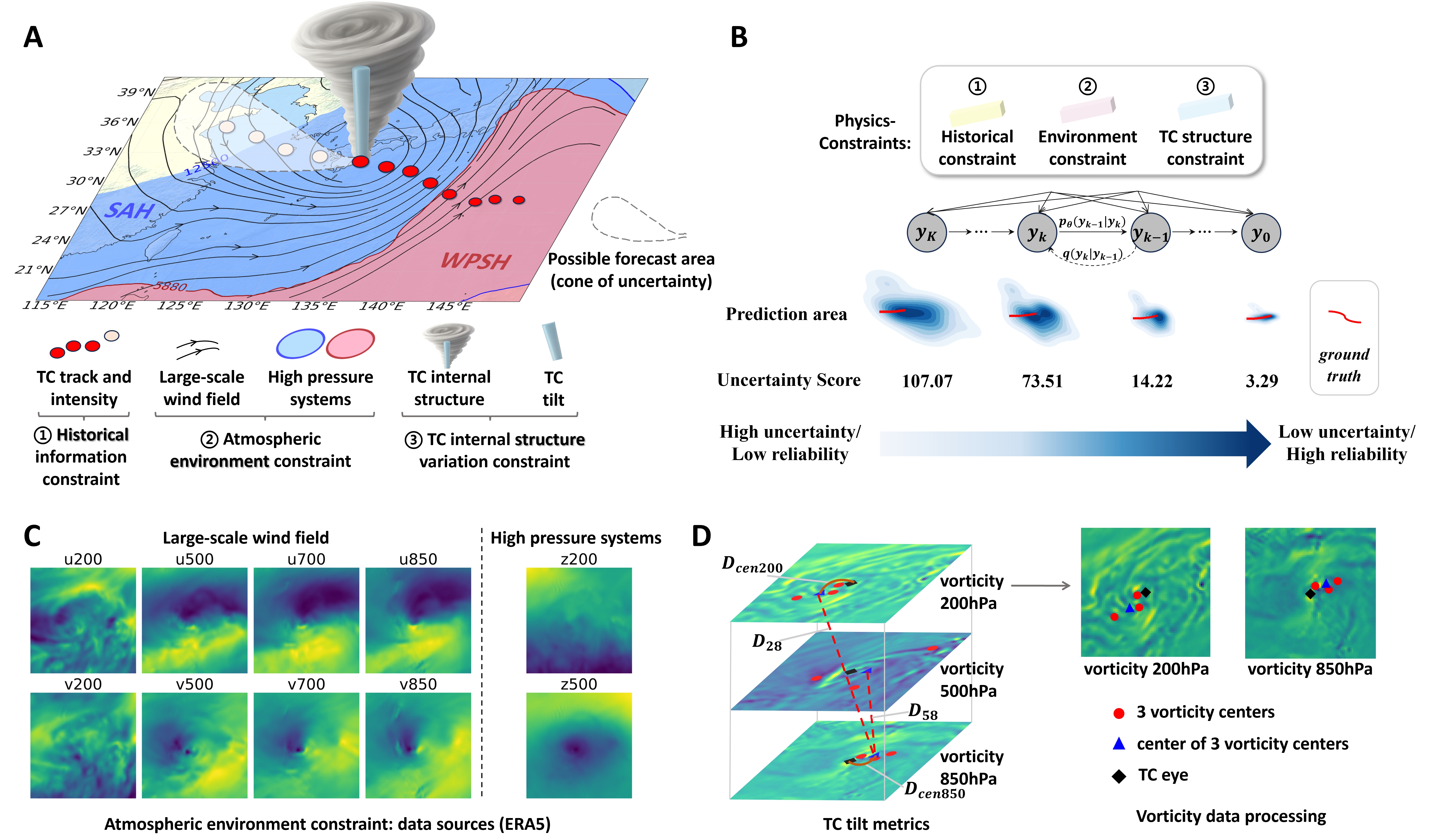}
	\caption{\textbf{Physics-constraints for TC forecasting.} (\textbf{A}) Overall schematic: meteorological factors influencing the development of TCs. (\textbf{B}) Proposed physics-constraints govern the generation process and achieve a lower uncertainty score with the smallest prediction area. (\textbf{C}) The data composition of atmospheric environment constraint. The large-scale wind field consists of the zonal \(u\) and meridional wind \(v\) components from 200 hPa to 850 hPa. High-pressure systems are composed of the South Asian High (SAH, \( z200 \)) and the Western Pacific Subtropical High (WPSH, \( z500 \)), where \( z \) denotes geopotential height. (\textbf{D}) TC tilt metrics: calculated from vorticity data at three pressure levels. See details in Materials and Methods. 
	}
	\label{fig:3constraints}
\end{figure}

However, the diversity reflected in these ensemble members—while beneficial for representing uncertainty—can lead to overly dispersed outputs, including multiple plausible but divergent tracks, thereby compromising the precision required for disaster prevention and emergency response. 
To address this, we introduce \textbf{physics-constraints} into the generative process. These constraints regulate the stochasticity of the model, enabling a more balanced trade-off between diversity and accuracy. The constraint conditions are formally defined and illustrated in Figure~\ref{fig:3constraints}.

\begin{figure}[!htb]    
	\centering 
	\begin{tabular}{c}
		\includegraphics[width=1\textwidth]{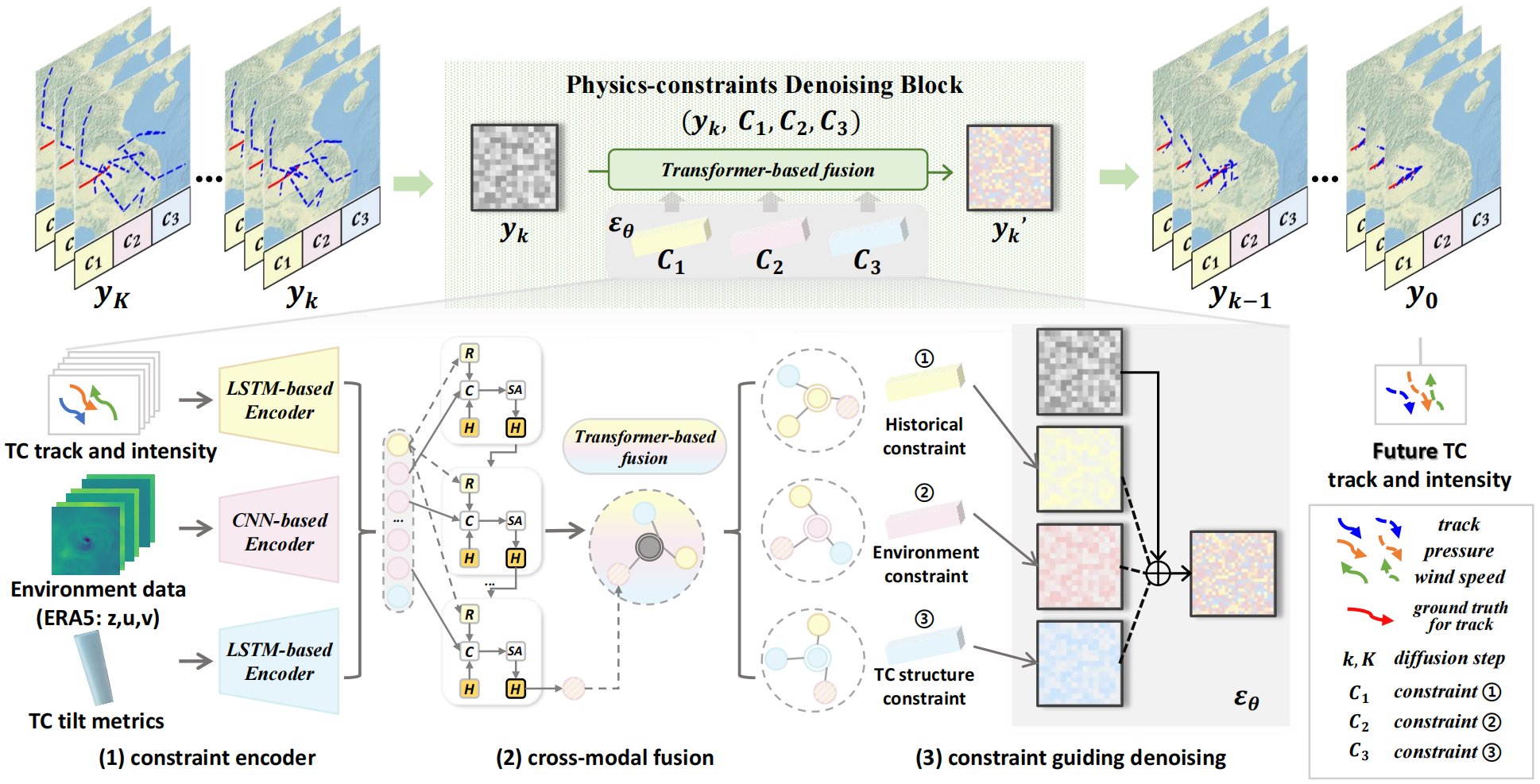}\\
	\end{tabular}
	\caption{\textbf{Framework of Tianmu-TC.} Tianmu-TC consists of three key components: \textbf{(1) Constraint Encoder} for preliminary constraint encoding, \textbf{(2) Cross-Modal Fusion} module to establish correlations between constraints, and \textbf{(3) Constraint guiding denoising module} to perform the final denoising process. 
	The $y_K \sim y_0$ represents the TC prediction results under different diffusion steps $k$. 
	To facilitate clear visualization, only the predicted track is displayed here, whereas wind speed and pressure predictions are simultaneously generated.
	From $y_K$ to $y_0$, the blue dashed line gradually converges from noise to near ground truth. The single-step denoising process from \( y_k \) to \( y_{k-1} \) is implemented by the \textbf{Physics-constraints Denoising Block}, which operates under the guidance of physics- constraints. } \label{fig:framework}
\end{figure}

Firstly, one common constraint in prediction tasks is the historical values of the target attribute. In TC forecasting, this typically includes the TC's past track and intensity, which we refer to as the \textbf{historical information constraint}.

Secondly, the generation and evolution of a TC are influenced by large-scale atmospheric environment. To account for this, we introduce \textbf{the atmospheric environment constraint}, which incorporate the major large-scale circulation systems influencing TC activity include the South Asia High (SAH) in the upper troposphere, the WP Subtropical High (WPSH), and the monsoon trough~\cite{wang2021impacts,wu2012possible,sun2015dependence}. 

Finally, beyond environmental forcing, TC attributes are also impacted by internal structural changes, which are often overlooked in prior work. Therefore, we designed the \textbf{internal structure variation constraint}, explicitly incorporating structural changes into the prediction process to control the diversity of generated outputs.

The three constraints above are collectively referred to as the \textbf{Physics-driven Meteorological Factor Constraints, namely, Physics-constraints}. Based on these, we introduce \textbf{Tianmu-TC}, a physics-constrained generative diffusion framework for global tropical cyclone forecasting, as shown in Figure~\ref{fig:framework}. By integrating Physics-constraints, Tianmu-TC transforms track and intensity prediction into a conditional generative process, where the initially chaotic generation results progressively become clearer and more accurate.
Our approach offers a new paradigm for efficient and robust global TC forecasting, highlighting the practicality and potential of generative modeling as a forecasting approach.

Previous efforts to apply artificial intelligence to TC forecasting have largely focused on deterministic prediction by learning patterns from historical data, while often lacking explicit modeling of forecast uncertainty, as exemplified by classical spatiotemporal sequence models such as  RNNs~\cite{moradi2016sparse,alemany2019predicting}, GRUs~\cite{cho2014learning}, Bi-GRU~\cite{song2022novel}, LSTMs~\cite{gao2018nowcasting}, and ConvLSTMs~\cite{kim2019deep}.
More recent work incorporates multi-modal inputs—including historical TC attributes and gridded reanalysis variables—to better represent the surrounding atmospheric environment~\cite{geng2023spatio,wu2021tropical,pan2019tropical,huang2022mmstn,huang2023mgtcf,wang2025global,huang2025benchmark,zhang2025tcdiffuser}.
In parallel, large-scale meteorological foundation models~\cite{bodnar2025foundation} such as Pangu~\cite{bi2023accurate} and Fengwu~\cite{chen2025operational}, together with recent probabilistic or ensemble AI weather models such as GenCast~\cite{price2025probabilistic} and NeuralGCM~\cite{kochkov2024neural}, have achieved impressive skill by directly forecasting global ERA5 reanalysis fields. Although these models can indirectly support TC forecasting through downstream tracking, they are not specifically optimized for TC-focused tasks. This often leads to underestimation of intensity, less compact ensemble distributions, and suboptimal performance in rare or extreme scenarios. Additionally, their dependence on globally scaled datasets and large model sizes incurs substantial training and inference costs, making them less practical for real-time or resource-constrained forecasting settings. 

In summary, existing approaches struggle to balance forecasting accuracy, computational efficiency, and uncertainty reduction, and often underperform under complex or rare meteorological conditions, such as rapid changes in tropical cyclone intensity or track.
Our goal is to enhance prediction reliability by effectively reducing uncertainty, without incurring additional computational cost.

\begin{figure}[htbp]    
	\centering 
	\begin{tabular}{c}
		\includegraphics[width=1.0\textwidth]{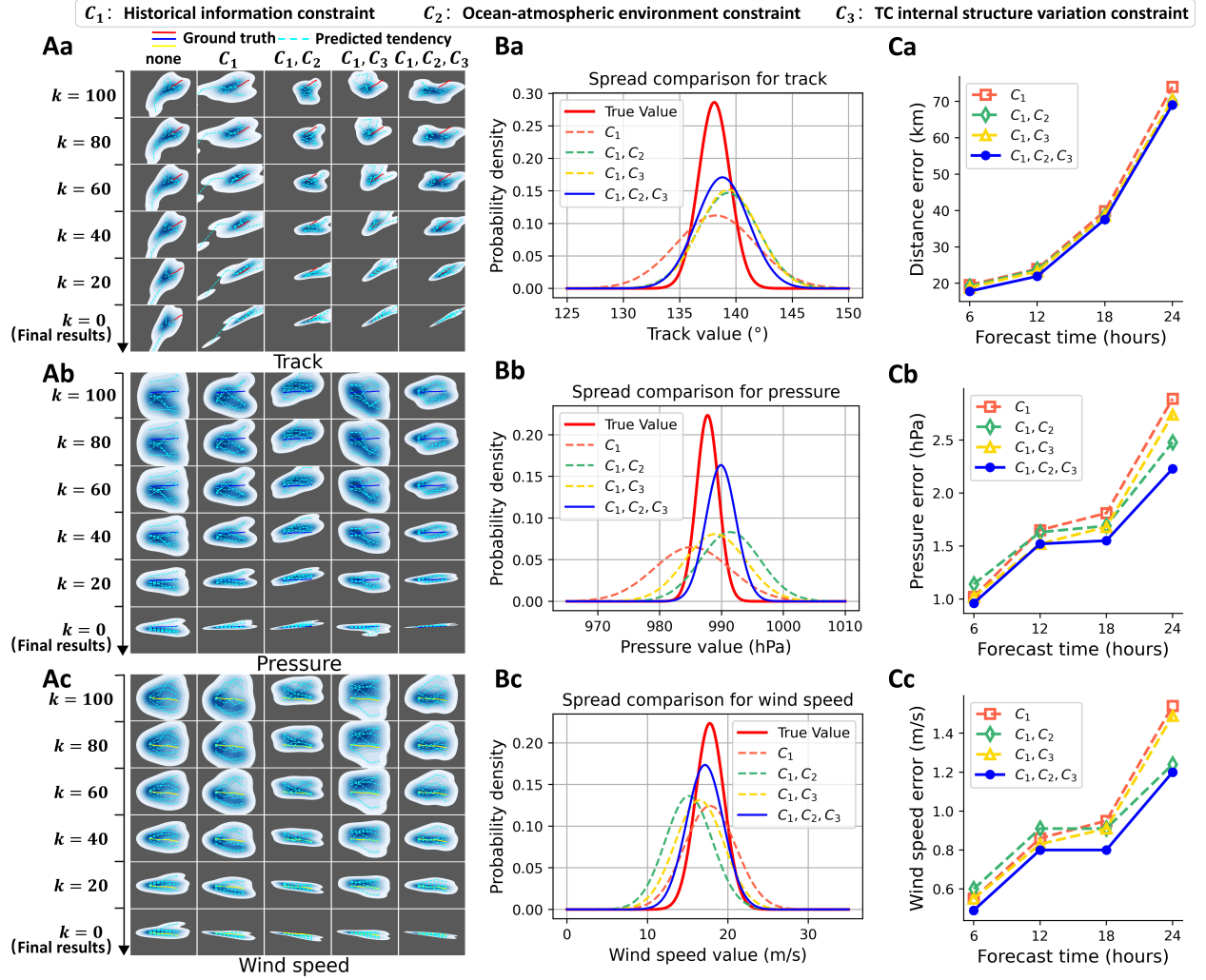}\\ 
	\end{tabular}
	\caption{\textbf{Results under different constraints.}
	(\textbf{A}) Generation results of Typhoon Krosa at 00:00 UTC on August 16, 2019, under five constraint settings with decreasing diffusion steps $k$ (from 100 to 0). Shaded regions represent uncertainty. Applying all constraints ($C_1$, $C_2$, and $C_3$; rightmost column) results in the most concentrated predictions with the lowest uncertainty. (\textbf{B}) For the Typhoon Krosa at 00:00 UTC on August 16, 2019, corresponding to (A), the mean and variance of the predictions are visualized as spread distributions. Smaller variance leads to a steeper curve, indicating lower prediction uncertainty. (\textbf{C}) Average prediction errors across 146 TCs from 2017 to 2023. The rows correspond to (\textbf{a}) track, (\textbf{b}) pressure, and (\textbf{c}) wind speed.
	}\label{fig:c1c2c3}
\end{figure}

\section*{Results}\label{sec2} 

Tianmu-TC takes 48 hours of TC historical data as input and generates forecasts for the next 24 or 72 hours. The input includes three core components: \ding{172} historical TC attributes from the International Best Track Archive for Climate Stewardship (IBTrACS~\cite{Knapp2010ibtracs}), \ding{173} ERA5-derived environmental fields representing large-scale atmospheric conditions, and \ding{174} TC tilt metrics extracted from vorticity fields to capture internal structural dynamics. In this study, the IBTrACS best track records are adopted as the ground truth in all evaluations, following standard practice in deep learning-based forecasting.

To rigorously assess generalizability, we evaluate Tianmu-TC on a global test set spanning all ocean basins from 2017 to 2023. This design ensures coverage of a wide range of TC scenarios, including varying intensities, tracks, and environmental conditions. Notably, to avoid excessive training time and computational cost, the model is trained exclusively on Western North Pacific (WP) TCs from 1950 to 2016. This highlights the model’s ability to generalize beyond its training domain, demonstrating robust performance across unseen basins despite being trained regionally. The time resolution is 6 hours. For fair comparison, following TCN$_{M}$~\cite{huang2025benchmark}, the sampling number was set to 6, meaning that the model generates 6 possible tendencies. Further details on the model architecture, training setup, and datasets are provided in the Materials and Methods section.

Tianmu-TC reduces forecast uncertainty through the incorporation of physics-driven constraints, which consists of the historical information constraint ($C_1$), the atmospheric environment constraint ($C_2$), and the TC internal structure variation constraint ($C_3$). We compare its performance against authoritative numerical weather prediction systems, deep learning–based methods, and deterministic and ensemble large-scale meteorological foundation models across global test regions. Additionally, we examine the model’s effectiveness under challenging scenarios such as data sparsity, anomaly tracks, landfall events, \textit{rapid intensification} and \textit{rapid weakening}.

The results demonstrate that Tianmu-TC achieves a favorable balance between predictive accuracy and controlled output diversity, while offering strong robustness and fast inference. Together with its compact and accurate multi-trend forecasts, these findings suggest that generative AI, when properly constrained by physical principles, constitutes a reliable and effective approach for advancing global TC forecasting.

\subsection*{Physics-constraints reduce prediction uncertainty}

A central challenge in TC forecasting is how to effectively reduce predictive uncertainty. Uncertainty arises naturally from the chaotic nature of the atmosphere and can be further amplified by model imperfections, making it essential to evaluate whether generated ensembles can provide not only accurate predictions but also reliable distributions. To address this, we investigate the role of incorporating physics-constraints into the generative process. We evaluate their effectiveness by comparing generation results under five settings: unconstrained generation (none of the constraints), generation with partial constraints ($C_1$, $C_1\&C_2$, $C_1\&C_3$), and generation with full physics-constraints ($C_1\&C_2\&C_3$). During reverse diffusion, the forecasts evolve from broad, highly uncertain distributions at diffusion step $k=100$ to concentrated predictions near the ground truth at $k=0$. 

These evaluations leads to three key observations:  

1. Full physics-constraints yield the lowest uncertainty with the most concentrated prediction distributions. When all three constraints are applied, Tianmu-TC generates ensembles that converge tightly around the ground truth for both track and intensity. As shown in Figure~\ref{fig:c1c2c3}Aa–Ac, the final outcomes ($k=0$) under full constraints form compact clusters, whereas the unconstrained case produces substantially larger deviations—track errors are about six times and intensity errors about three times larger. This aspect is further illustrated by the probability density curves in Figure~\ref{fig:c1c2c3}Ba–Bc, where the fully constrained results exhibit the steepest probability density curves, with data points highly concentrated, indicating concentrated forecasts and substantially reduced uncertainty. The Figure~\ref{fig:uq_score} presents the uncertainty score corresponding to each predicted region, calculated as the MSE loss between the predictions and the ground truth. This further supports the aforementioned conclusion.

2. Physics-constraints enable the forecast distribution to better align with the ground-truth distribution, thereby improving both local accuracy and global consistency. As shown in Figure~\ref{fig:c1c2c3}Ba–Bc, the probability density curves of Tianmu-TC exhibit peak values and shapes closest to those of the ground truth, indicating that predictions are not only accurate at specific points but also capture the broader structural trends of tropical cyclones. This distributional alignment highlights the model’s ability to reproduce realistic ensemble behavior, extending beyond reduced spread to better capturing the statistical consistency of forecasts with the ground truth.

3. Physics-constraints reduce average prediction errors. As shown in Figure~\ref{fig:c1c2c3}Ca–Cc, incorporating all three constraints lowers the average errors across track, pressure, and intensity, confirming the higher accuracy of Tianmu-TC compared with alternative settings. Specifically for intensity prediction, the $C_3$ (TC internal structure variation constraint) enhances forecasts by encoding vertical structural changes within the TC, consistent with the well-established relationship between cyclone structure and intensity evolution~\cite{wang2004current}. This agreement reinforces the physical interpretability of the model.

The effectiveness of physics-constraints in reducing prediction uncertainty is consistently reflected in visualization patterns, distribution convergence, and average error metrics, thereby resulting in improved forecasting performance.

\begin{figure}[!htb]    
	\centering 
	\begin{tabular}{c}
		\includegraphics[width=1\textwidth]{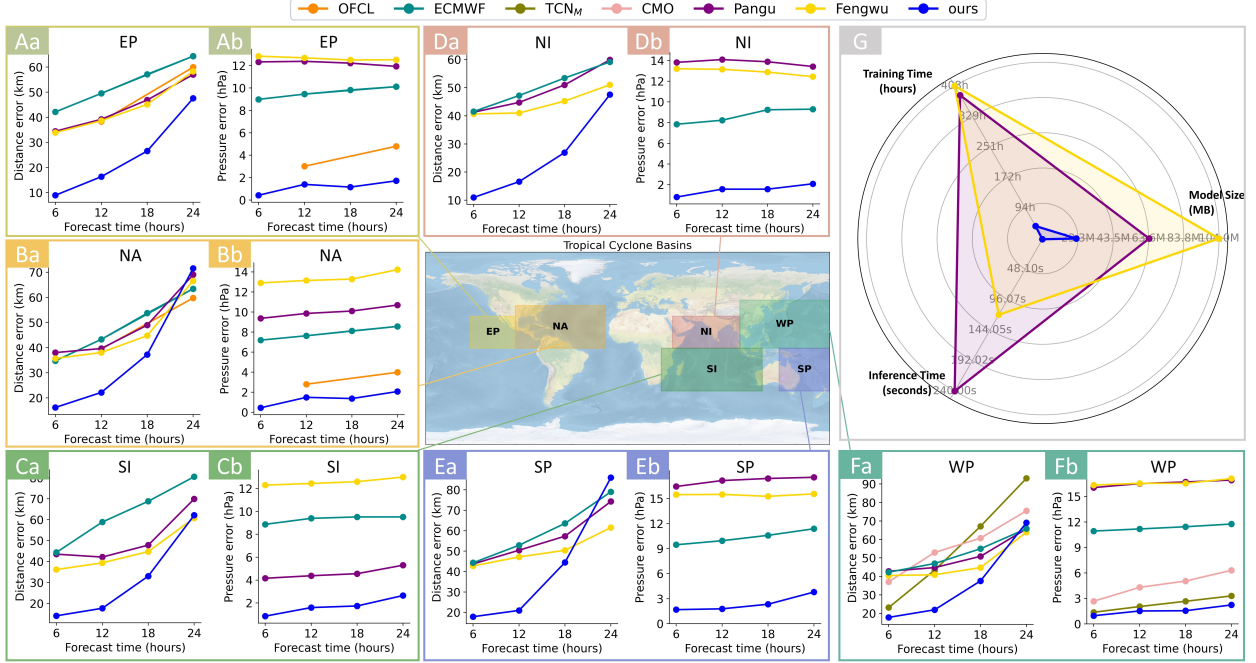}\\
	\end{tabular}
	\caption{\textbf{Comparison of average forecast errors with large-scale meteorological models and the authoritative NWP system during the test period from 2017 to 2023.} The testing regions cover six ocean basins: (\textbf{A}) Eastern North Pacific (EP), (\textbf{B}) North Atlantic (NA), (\textbf{C}) North Indian (NI), (\textbf{D}) South Indian (SI), (\textbf{E}) South Pacific (SP), and (\textbf{F}) Western North Pacific (WP). (\textbf{G}) Comparison of computational efficiency metrics with large-scale meteorological models.}
	 \label{fig:global2}
\end{figure}

\begin{figure}[!htb]    
	\centering 
	\begin{tabular}{c}
		\includegraphics[width=1\textwidth]{framework/global-visual.png}\\
	\end{tabular}
	\caption{\textbf{Visualization of track predictions for representative TCs across six ocean basins.} Each basin presents forecasts for a specific TC at three consecutive initialization times: (\textbf{A}) EP (Jimena), (\textbf{B}) NA (Rene), (\textbf{C}) NI (Ockhi), (\textbf{D}) SI (Enawo), (\textbf{E}) SP (Gita), and (\textbf{F}) WP (Sonca).} \label{fig:global_visual}
\end{figure}

\subsection*{Generalizability on global ocean areas} 

Tropical cyclones occur across multiple ocean basins, making cross-regional generalization a critical requirement for forecasting models. Although Tianmu-TC is trained exclusively on Western North Pacific data, we evaluate its performance on global TC datasets spanning six basins to assess its transferability to real-world applications. As shown in Figure~\ref{fig:global2}, Tianmu-TC outperforms large meteorological models such as Pangu~\cite{bi2023accurate} and Fengwu~\cite{chen2025operational} in most regions, while achieving comparable overall performance. In addition, Figure~\ref{fig:global2}G shows that Tianmu-TC achieves an eightfold reduction in training time, a tenfold reduction in inference time, and a threefold reduction in model size. Two main conclusions emerge: 

\begin{enumerate}
	\item \textbf{Efficiency and performance superiority.} Despite its lower training costs, smaller model size, and faster inference compared with large meteorological models, Tianmu-TC achieves comparable track prediction and significantly better intensity prediction. Notably, it significantly outperforms ECMWF (the ECMWF global model \cite{ecmwf2011ifs}) in the EP, NI, and SI basins (Figure~\ref{fig:global2}A,C,D). In the NA, SP, and WP basins (Figure~\ref{fig:global2}B,E,F), it delivers superior intensity prediction performance across all metrics and competitive track accuracy within 18 hours. Moreover, Tianmu-TC runs on a single NVIDIA A6000 GPU rather than supercomputing infrastructure, highlighting that deep learning can surpass operational NWP while requiring substantially fewer computational resources. 
	\item \textbf{Cross-regional scalability.} Although trained only on WP data, Tianmu-TC generalizes effectively to the other five basins (Figure~\ref{fig:global2}A–E), capturing distinct TC dynamics in both hemispheres. This demonstrates robust adaptability to diverse environments and strong potential for global TC forecasting.
\end{enumerate}
\begin{table}[htb]
	\centering
	\caption{\textbf{Results on special TC scenarios on global area from 2017 to 2021.} 
		Samples is the number of cases. Lower values indicate better performance. ``Anomaly'' consists of circular movement, turning on the ocean, and V-shaped tracks. ``\textit{RI}'' denotes \textit{Rapid Intensification}, and ``\textit{RW}'' denotes \textit{Rapid Weakening} of TCs.}
	\label{tab:special_scenarios}
	\footnotesize
	\setlength{\tabcolsep}{2.6pt}
	\begin{tabular}{lllcccccc}
		\hline
		\multirow{2}{*}{Category} & \multirow{2}{*}{Scenario (Unit)} & \multirow{2}{*}{Methods} & \multirow{2}{*}{Samples} & \multicolumn{4}{c}{Forecast Horizon} \\
		\cline{5-8}
		&  &  &  & +6h & +12h & +18h & +24h \\
		\hline
		\multirow{8}{*}{\textbf{Track}} 
		& \multirow{4}{*}{Anomaly (km)} 
		& Pangu  & 582 & 44.13 & 55.90 & 72.75 & 104.26 \\
		& & Fengwu & 582 & 44.19 & 49.33 & 65.89 & 95.62 \\
		& & ECMWF   & 582 & 40.32 & 50.26 & 61.82 & 78.11 \\
		& & Ours   & 582 & \textbf{16.90} & \textbf{17.28} & \textbf{34.74} & \textbf{70.38} \\
		\cline{2-8}
		& \multirow{4}{*}{Landfall (km)} 
		& Pangu  & 304 & 47.86 & 62.09 & 87.55 & 125.50 \\
		& & Fengwu & 304 & 43.96 & 52.80 & 65.19 & 83.17 \\
		& & ECMWF   & 304 & 35.71 & 47.24 & 57.08 & 76.26 \\
		& & Ours   & 304 & \textbf{16.98} & \textbf{20.90} & \textbf{36.61} & \textbf{66.26} \\
		\hline
		\multirow{8}{*}{\textbf{Intensity}} 
		& \multirow{4}{*}{RI (m/s)} 
		& Pangu  & 231 & 22.89 & 31.68 & 41.30 & 49.32 \\
		& & Fengwu & 231 & 21.89 & 29.85 & 38.88 & 46.17 \\
		& & ECMWF  & 231 & 17.41 & 18.39 & 24.32 & 28.78 \\
		& & Ours   & 231 &  \textbf{1.35} &  \textbf{1.29} &  \textbf{4.41} &  \textbf{7.59} \\
		\cline{2-8}
		& \multirow{4}{*}{RW (m/s)} 
		& Pangu  & 320 & 28.69 & 22.05 & 16.07 & 12.01 \\
		& & Fengwu & 320 & 27.87 & 21.24 & 14.94 & 10.49 \\
		& & ECMWF   & 320 & 13.78 & 10.73 &  8.64 & 26.40 \\
		& & Ours   & 320 &  \textbf{2.69} &  \textbf{1.63} &  \textbf{3.79} &  \textbf{5.74} \\
		\hline
	\end{tabular}
\end{table}

\begin{figure}[!htb]
	\centering 
	\begin{tabular}{c}
		\includegraphics[width=0.98\textwidth]{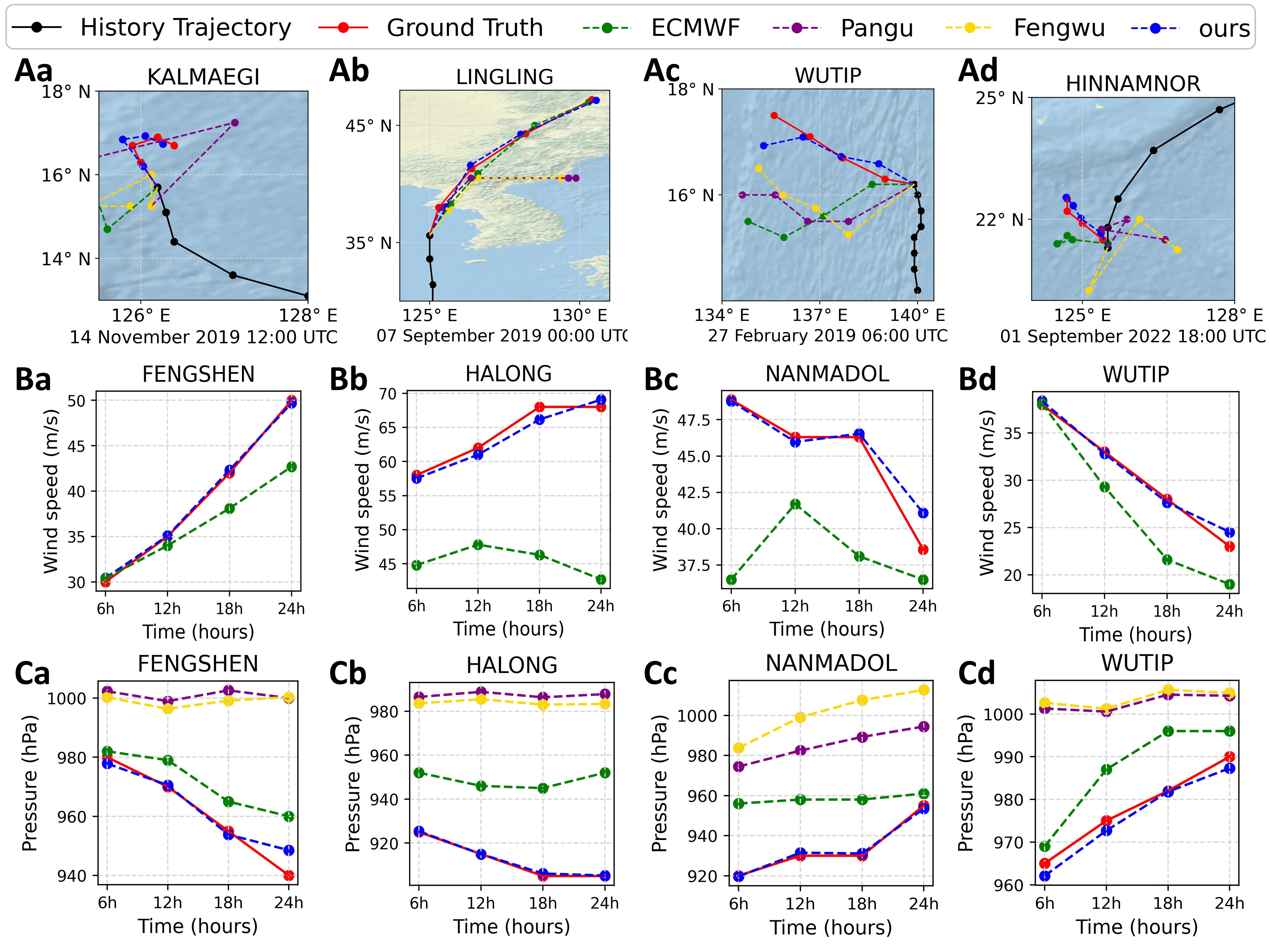}\\
	\end{tabular}	
	\caption{\textbf{Visualization of complex prediction scenarios.} (\textbf{A}) Track prediction, (\textbf{B}) wind speed, and (\textbf{C}) pressure. For (A), four challenging track cases are illustrated: (\textbf{Aa}) circular track, (\textbf{Ab}) post-landfall prediction, (\textbf{Ac}) turning on the ocean, and (\textbf{Ad}) V-shaped movement. (\textbf{Ba})–(\textbf{Bb}): \textit{rapid intensification}; (\textbf{Bc})–(\textbf{Bd}): \textit{rapid weakening}. \textit{Rapid intensification} is defined as a tropical cyclone's wind speed increasing by more than 30 knots (15.43 m/s) within 24 hours, whereas \textit{rapid weakening} refers to the opposite process. (\textbf{C}) displays the corresponding pressure prediction results for (B).}
	\label{fig:special}
\end{figure}

For broader comparison, we also benchmarked Tianmu-TC against authoritative forecasts provided by local meteorological agencies in Figure~\ref{fig:global2} and Figure~\ref{fig:supple-compare}. The official forecast issued by the National Hurricane Center (OFCL)~\cite{nhcverify} was assessed in the EP and NA basins, while the NWP system of the China Central Meteorological Observatory (CMO)~\cite{CMO} was evaluated in the WP basin. As shown in Figure~\ref{fig:global2}A, Tianmu-TC significantly outperforms OFCL in the EP basin. Notably, Tianmu-TC surpasses for the first time the authoritative NWP model operated by the CMO across all metrics in the WP basin (Figure~\ref{fig:global2}F).

Most existing studies focus on the Western North Pacific, where tropical cyclones occur most frequently. Therefore, we further compared Tianmu-TC with deep learning–based approaches in the WP basin (Figure~\ref{fig:global_wp}). These include single-task baselines using only historical track or intensity information (DLM~\cite{pan2019tropical} and TCIF-fusion~\cite{wang2024tropical}) and multi-task models (SGAN~\cite{gupta2018social}, GBRNN~\cite{alemany2019predicting}, MMSTN~\cite{huang2022mmstn}, TCN$_{M}$~\cite{huang2025benchmark}, TC-Diffuser~\cite{zhang2025tcdiffuser} and TAM-CL~\cite{li2024tropical}). Tianmu-TC consistently surpasses all of these methods across evaluation metrics, highlighting its advantages in both track and intensity prediction.

In addition to aggregated performance metrics, Figure~\ref{fig:global_visual} presents qualitative visualizations of track predictions across the six global ocean basins. For each basin, we selected a representative tropical cyclone and visualized its predicted tracks at three consecutive forecast timestamps. This enables the comparison of temporal continuity and accuracy of predictions among different models. Tianmu-TC shows consistent alignment with ground truth tracks over time, capturing not only general movement direction but also subtle tracks turns, demonstrating its reliability and temporal coherence in global-scale forecasting.

Overall, these experiments underscore the practical potential of Tianmu-TC: it not only reduces training costs but also demonstrates exceptional competitiveness against both operational forecasts and state-of-the-art deep learning approaches. 

\subsection*{Comparison with ensemble meteorological AI model: GenCast \& NeuralGCM}

To compare with existing multi-trend forecasting methods, GenCast and NeuralGCM were further selected as representative global machine-learning weather forecasting models for supplementary evaluation. These models were chosen because they provide both deterministic and ensemble forecasts, making them suitable references for assessing the effectiveness and competitiveness of Tianmu-TC in multi-trend tropical cyclone forecasting. To ensure reproducibility, the publicly archived gridded forecast data for the year 2020 were used, and tropical cyclone centers and intensity diagnostics were derived directly from the original forecast fields. Details of the data sources, selected forecast variables, tracking rules, and intensity diagnosis are provided in Supplementary Methods (Sections~\ref{supple:neuralgcm} and~\ref{supple:gencast}).

This comparison is conducted only for the year 2020 because publicly released GenCast and NeuralGCM forecasts are only provided for this year, and this avoids potential temporal overlap with the training period of GenCast. Since these forecasts are provided at 12 h intervals, the evaluation is performed at the shared forecast horizons of +12 h and +24 h. In addition, because MSLP is not available in the original NeuralGCM configuration used here, only track prediction results are reported for NeuralGCM. 

\begin{table}[htb]
	\centering
	\caption{\textbf{Comparison with deterministic NeuralGCM and GenCast forecasts.}
		Results are reported on test samples for the year 2020 at the overlapping forecast horizons of +12 h and +24 h.
		Track error, pressure error, and wind speed error are measured in km, hPa, and m/s, respectively.
		Lower values indicate better performance. In “Inference time”, s denotes seconds.}
	\label{tab:gencast_deterministic}
	\footnotesize
	\setlength{\tabcolsep}{3.0pt}
	\begin{tabular}{ccccccccccccc}
		\hline
		\multirow{2}{*}{Model} & \multirow{2}{*}{Samples}
		& \multicolumn{3}{c}{Track Error (km)}
		& \multicolumn{3}{c}{Pressure Error (hPa)}
		& \multicolumn{3}{c}{Wind Speed Error (m/s)} & \multicolumn{1}{c}{\multirow{2}{*}{\makecell{Training\\ time}}}&\multicolumn{1}{c}{\multirow{2}{*}{\makecell{Inference\\ time}}}\\
		\cline{3-11}
		& & +12h & +24h & Overall
		& +12h & +24h & Overall
		& +12h & +24h & Overall & & \\
		\hline
		NeuralGCM &684 &67.52&83.28&75.40&/&/&/&/&/&/&3 weeks &11.9 s\\
		GenCast & 684
		& 43.25 & \textbf{62.00} & 52.63
		& 10.71 & 10.75 & 10.73
		& 10.10 & 9.93 & 10.02&5 days &32 s \\
		Tianmu-TC & 684
		& \textbf{17.52} & 66.37 & \textbf{41.95}
		& \textbf{0.39} & \textbf{1.46} & \textbf{0.92}
		& \textbf{0.72} & \textbf{2.11} & \textbf{1.41} &2 days &1.23 s \\
		\hline
	\end{tabular}
\end{table}

Table~\ref{tab:gencast_deterministic} shows the \textbf{deterministic forecast comparison}. Tianmu-TC clearly outperforms NeuralGCM in track prediction. This gap may be partly associated with the relatively coarse \((0.7^\circ)\) spatial resolution of the public deterministic NeuralGCM forecasts, which limits the precision of grid-based TC center localization. In contrast, Tianmu-TC is trained with high-precision best-track coordinates and is specifically optimized for short-term TC track prediction.
Compared with GenCast, Tianmu-TC achieves lower errors for the +12 h track, the overall track, and all intensity-related metrics, while its +24 h track error is only slightly higher. These results indicate that Tianmu-TC provides more accurate short-term tropical cyclone intensity estimates and comparable track prediction skill relative to GenCast.
This comparison is also informative from the perspective of computational cost. As a global medium-range ensemble forecasting system, GenCast is built at a substantially larger scale: the original GenCast paper reports slightly more than 3.5 days of pretraining and under 1.5 days of 0.25° fine-tuning on 32 TPUv5 instances, while a single 15-day global forecast takes approximately 8 minutes on a Cloud TPUv5 device. Under a simple linear scaling by forecast length, this corresponds to approximately 32 s for a 24 h forecast. In contrast, Tianmu-TC is specifically designed for short-term multi-attribute tropical cyclone forecasting, requires only 1.23 s for prediction, and can be trained on a single A6000 GPU in approximately 2 days.
These results suggest that, with substantially lower computational cost, a tropical-cyclone-specific model still has clear advantages for short-term tropical cyclone track and intensity forecasting.

\begin{table*}[htb]
	\centering
	\caption{\textbf{Comparison with ensemble NeuralGCM and GenCast forecasts.}
		Results are reported on test samples for the year 2020 at the overlapping forecast horizons of +12 h and +24 h.
		The overall value is computed over the combined +12 h and +24 h samples.
		NeuralGCM-ENS, GenCast-ENS and Tianmu-TC-ENS denote the ensemble versions of NeuralGCM, GenCast and Tianmu-TC, respectively.
		For a fair comparison, all ensemble-based methods are evaluated using 50 members.}
	\label{tab:gencast_ngcm_ensemble}
	\footnotesize
	\setlength{\tabcolsep}{2.0pt}
	\resizebox{\textwidth}{!}{
		\begin{tabular}{llccccccccccccc}
			\hline
			\multirow{2}{*}{Task}
			& \multirow{2}{*}{Model}
			& \multirow{2}{*}{Samples}
			& \multicolumn{3}{c}{Mean Error}
			& \multicolumn{3}{c}{Std. Error}
			& \multicolumn{3}{c}{Spread}
			& \multicolumn{3}{c}{Best-of-50} \\
			\cline{4-15}
			& & 
			& +12h & +24h & Overall
			& +12h & +24h & Overall
			& +12h & +24h & Overall
			& +12h & +24h & Overall \\
			\hline
			
			\multirow{3}{*}{Track (km)}
			& NeuralGCM-ENS & 521/505
			& 91.75 & 126.87 & 109.04
			& 31.88 & 57.18 & 44.33
			& 95.02 & 145.47 & 119.85
			& 62.36 & 67.16 & 64.72 \\
			
			& GenCast-ENS & 521/505
			& 92.28 & 175.74 & 133.36
			& 51.91 & 104.05 & 77.57
			& 96.40 & 189.52 & 142.24
			& 24.76 & \textbf{39.86} & 32.19 \\
			
			& Tianmu-TC-ENS & 521/505
			& \textbf{28.40} & \textbf{109.86} & \textbf{68.49}
			& \textbf{8.76} & \textbf{34.16} & \textbf{21.26}
			& \textbf{22.33} & \textbf{85.29} & \textbf{53.32}
			& \textbf{10.42} & 40.94 & \textbf{25.44} \\
			\hline
			
			\multirow{3}{*}{Pressure (hPa)}
			& NeuralGCM-ENS & /
			& / & / & /
			& / & / & /
			& / & / & /
			& / & / & / \\
			
			& GenCast-ENS & 521/505
			& 9.72 & 10.05 & 9.88
			& 1.30 & 1.92 & 1.60
			& 1.34 & 2.05 & 1.69
			& 6.88 & 6.27 & 6.58 \\
			
			& Tianmu-TC-ENS & 521/505
			& \textbf{1.66} & \textbf{4.90} & \textbf{3.26}
			& \textbf{0.59} & \textbf{1.72} & \textbf{1.14}
			& \textbf{0.67} & \textbf{1.94} & \textbf{1.29}
			& \textbf{0.58} & \textbf{1.72} & \textbf{1.14} \\
			\hline
			
			\multirow{3}{*}{Wind Speed (m/s)}
			& NeuralGCM-ENS & /
			& / & / & /
			& / & / & /
			& / & / & /
			& / & / & / \\
			
			& GenCast-ENS & 521/505
			& 9.34 & 9.28 & 9.31
			& 1.14 & 1.52 & 1.33
			& 1.18 & 1.60 & 1.39
			& 6.72 & 6.15 & 6.44 \\
			
			& Tianmu-TC-ENS & 521/505
			& \textbf{1.01} & \textbf{3.16} & \textbf{2.07}
			& \textbf{0.36} & \textbf{1.06} & \textbf{0.70}
			& \textbf{0.42} & \textbf{1.21} & \textbf{0.81}
			& \textbf{0.34} & \textbf{1.22} & \textbf{0.77} \\
			\hline
		\end{tabular}
	}
\end{table*}

\begin{figure}[!htb]
	\centering 
	\begin{tabular}{c}
		\includegraphics[width=1\textwidth]{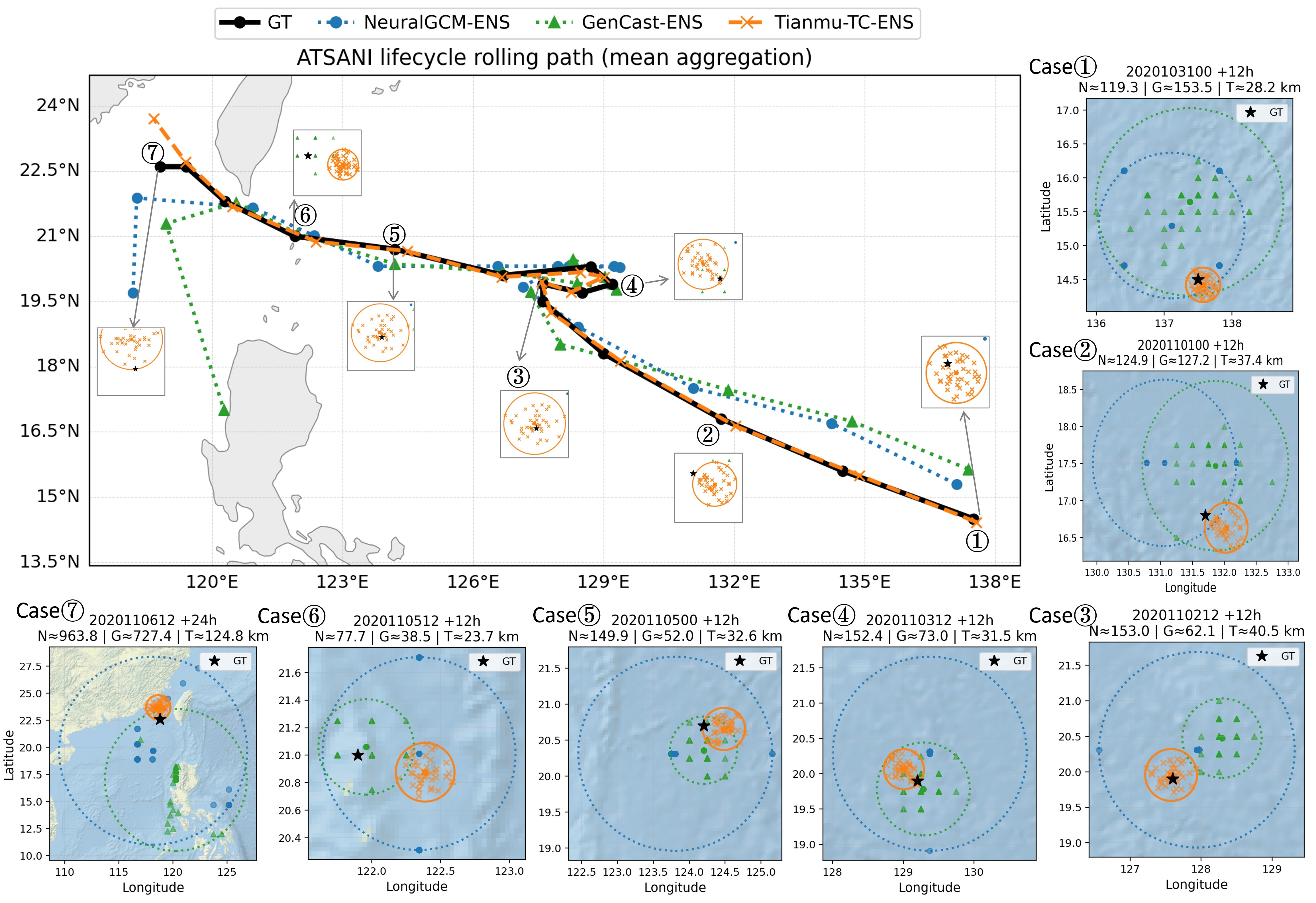}\\
	\end{tabular}	
	\caption{
		\textbf{Lifecycle visualization of multi-trend track forecasts for Typhoon ATSANI (WP, 2020).}
		The main panel shows the rolling 12-h forecast path over the TC lifecycle, where each method is represented by the mean track of 50 ensemble members (i.e., multiple possible tracks), and the black curve denotes the best-track positions. 
		The numbered panels show the spatial distributions of the 50 ensemble-predicted TC positions for representative cases selected from the main panel, covering early, middle, and late lifecycle stages as well as unusual looping-track predictions.
		In each numbered panel, circles indicate the spatial dispersion of the forecast members, and the values $N$, $G$, and $T$ denote the corresponding dispersion radii of NeuralGCM-ENS, GenCast-ENS, and Tianmu-TC-ENS, respectively. 
		The black star denotes the best-track position. 
		The selected cases reveal different performance regimes, including accurate and biased situations.
	}
	\label{fig:atsani_lifecycle_multitrend}
\end{figure}

For \textbf{multi-trend prediction}, a formulation analogous to ensemble forecasting in numerical weather prediction, Table~\ref{tab:gencast_ngcm_ensemble} evaluates 50 sampled trends from four complementary perspectives.
Mean Error and Std. Error denote the mean and standard deviation of the errors across the 50 ensemble members or sampled trends, respectively. 
Spread measures the dispersion of the 50 forecasts and is used as an uncertainty-related indicator of the forecast distribution. 
Specifically, for track prediction, Spread is computed as the 90th-percentile radius of the distances between the 50 ensemble-predicted TC positions and the ensemble center, while for pressure and wind speed it is computed from the dispersion of the 50 predicted scalar values.
Best-of-50 evaluates the error of the best member among the 50 forecasts, reflecting whether the forecast distribution contains a highly accurate candidate.

First, Tianmu-TC-ENS improves the accuracy and stability of multi-trend prediction. 
For track prediction, Tianmu-TC-ENS reduces the overall Mean Error from 109.04 km and 133.36 km to 68.49 km compared with NeuralGCM-ENS and GenCast-ENS, respectively. 
The overall Std. Error is also reduced from 44.33 km and 77.57 km to 21.26 km, indicating that the sampled trends are not only closer to the ground truth on average, but also more stable across different members. 
For intensity prediction, Tianmu-TC-ENS similarly reduces the overall Mean Error from 9.88 hPa to 3.26 hPa for pressure and from 9.31 m/s to 2.07 m/s for wind speed, further confirming its advantage in multi-attribute TC prediction.

Second, Tianmu-TC-ENS produces a more compact forecast distribution while preserving accurate candidates. 
For track prediction, the overall Spread is reduced from 119.85 km and 142.24 km to 53.32 km, showing that Tianmu-TC-ENS substantially suppresses unnecessary spatial dispersion among the 50 trends. 
For pressure and wind speed, Tianmu-TC-ENS also reduces the overall Spread from 1.69 hPa to 1.29 hPa and from 1.39 m/s to 0.81 m/s, respectively. 
Meanwhile, the Best-of-50 results show that this compactness does not come from a simple collapse of the forecast distribution: Tianmu-TC-ENS achieves the best overall Best-of-50 errors for track, pressure, and wind speed prediction, although GenCast-ENS is slightly better for the +24 h track Best-of-50 case. 

Overall, these results provide quantitative evidence that the physically constrained multi-trend formulation of Tianmu-TC-ENS can improve the accuracy of sampled TC evolution trends while reducing unnecessary predictive dispersion. 
This supports the central objective of Tianmu-TC-ENS, namely generating physically consistent and uncertainty-compact multi-trend forecasts for TC track and intensity prediction.

Figure~\ref{fig:atsani_lifecycle_multitrend} further evaluates the lifecycle behavior of multi-trend track forecasts for Typhoon ATSANI. Rather than focusing on a single initialization time, the figure examines forecasts throughout the entire TC lifecycle. ATSANI is selected because it provides diverse and challenging forecast scenarios, including unusual looping-track predictions, a pronounced track deflection near southern Taiwan, and subsequent evolution close to coastal and mountainous regions. These scenarios enable forecast stability to be evaluated under varying environmental and topographic conditions. Specifically, each method is represented by the mean track of 50 ensemble members, allowing a direct and fair comparison of lifecycle-scale track evolution, the dispersion of 50 ensemble-predicted TC positions, and the ability of the forecast distribution to cover the best-track position. As shown in the predicted positions panels of Figure~\ref{fig:atsani_lifecycle_multitrend}, the values $N$, $G$, and $T$ denote the dispersion radii of NeuralGCM-ENS, GenCast-ENS, and Tianmu-TC-ENS, respectively, with smaller values indicating lower spatial spread among the 50 forecasts. Tianmu-TC-ENS produces a more compact multi-trend forecast distribution, indicating lower forecast uncertainty while still covering the best-track position in most selected cases. In cases\ding{172}--\ding{176}, the orange Tianmu-TC-ENS predicted positions are concentrated within a much smaller circle than the blue NeuralGCM-ENS and green GenCast-ENS predicted positions, and the best-track position is generally located inside or close to the Tianmu-TC-ENS distribution. For example, Tianmu-TC-ENS shows notably smaller dispersion in cases\ding{172}, \ding{173}, and \ding{175}, where the orange predicted positions remain tightly clustered around the best-track position. Meanwhile, NeuralGCM-ENS and GenCast-ENS usually exhibit larger circles, indicating stronger ensemble spread. This larger spread can also be beneficial, as it enables the global ensemble forecasts to cover the best-track position in several cases, although at the cost of higher uncertainty. These results suggest that Tianmu-TC-ENS provides a more concentrated multi-trend forecast while maintaining useful best-track coverage in most stages of the ATSANI lifecycle.

The selected cases also reveal several limitations. In case\ding{177}, Tianmu-TC-ENS shows the largest positional bias among the three methods, whereas GenCast-ENS places its members closer to the best-track position. This error may be related to the relatively fast storm displacement in the preceding stages, which increases the difficulty of short-term track extrapolation. In case\ding{178}, all three methods exhibit large errors, while NeuralGCM-ENS and GenCast-ENS show particularly large dispersion. This is likely associated with the late-stage land interaction and the rapidly changing storm environment near landfall. In addition, although GenCast-ENS usually benefits from its larger spread, it fails to cover the best-track position in case\ding{174}. These results indicate that Tianmu-TC-ENS can substantially reduce excessive ensemble dispersion and provide stable multi-trend forecasts in most situations. However, achieving both low dispersion and reliable best-track coverage remains an important direction for future improvement.

\subsection*{Robust forecasting of abrupt track changes and RI/RW events}
To evaluate the performance of Tianmu-TC under challenging conditions, we examined its predictions for abrupt track changes and rapid intensity changes, which remain particularly difficult to forecast~\cite{lee2016rapid}. As shown in Table~\ref{tab:special_scenarios}, across abrupt track changes (denoted as \textbf{anomaly} in the table), landfall, \textit{rapid intensification (\textbf{RI})}, and \textit{rapid weakening (\textbf{RW})} cases, Tianmu-TC consistently achieves the lowest forecast errors compared with the large meteorological models and operational NWP models (Pangu, Fengwu, ECMWF), demonstrating superior robustness under complex conditions. \textit{Anomaly} includes circular movement, turning on the ocean, and V-shaped tracks; \textit{landfall} refers to samples in which landfall occurs during at least half of the forecast period (24 hours). For track anomalies, the errors of Tianmu-TC are reduced by up to a factor of six relative to Pangu and by a factor of three relative to Fengwu. In landfall prediction, it substantially improves accuracy over ECMWF and achieves more than a twofold error reduction compared with Pangu. For intensity variation, Tianmu-TC reduces RI errors to less than one-tenth of those from operational NWP models and also achieves the best performance under RW cases. Since ECMWF dataset is limited, we additionally report comparisons on larger samples (aligned with Tianmu-TC, Pangu, Fengwu, see Table~\ref{tab:special_scenarios_extend}). The performance trends remain consistent, further validating the robustness of Tianmu-TC.

Besides, the corresponding visualization results are shown in Figure~\ref{fig:special}. For tracks, four representative scenarios were analyzed: circular movement, post-landfall prediction, turning on the ocean, and V-shaped tracks. For intensity, we focused on cases of \textit{rapid intensification} and \textit{weakening}. It can be observed that Tianmu-TC effectively captures the dynamics and transitions in these scenarios, providing reliable track forecasts even under irregular movements and maintaining high accuracy in predicting rapid intensity changes. 

These results confirm that the proposed physics-constrained generative framework is capable of handling the most difficult TC scenarios, where both track and intensity variations remain highly uncertain in conventional approaches.

\subsection*{Accurate real-time prediction under data sparsity}\label{sec:real-time}
Given that the ERA5 reanalysis data used to encode the atmospheric environment constraint are not available in real time, we examined Tianmu-TC’s performance in cases where ERA5 data were missing during testing (see Table~\ref{tab:datamissing}). In such cases, critical environmental inputs—including wind fields ($u$, $v$) and geopotential height ($z$) representing high-pressure systems—were unavailable. 

By comparing Rows 9 and 12 in Table~\ref{tab:datamissing}, we find that the absence of ERA5 data during inference has negligible impact on prediction accuracy. Notably, Tianmu-TC maintains comparable performance using only the TC historical attributes from the previous 24 hours. 

This finding highlights the model's robustness under data sparsity, where only partial inputs are available, and confirms its practical potential as a real-time, end-to-end TC forecasting method, even in the absence of comprehensive environmental data.

\begin{table}[htb]
	\centering
	\caption{\textbf{Testing results under missing input conditions.} 
		The last row corresponds to the complete model (Tianmu-TC). $X_{his}$, $X_{ERA5}$, $X_{tilt}$ refer to TC historical attribute values, ERA5 data, and TC tilt values, respectively. $+6h$ to $+24h$ indicate prediction horizons in hours. Lower values represent better performance.}
	\label{tab:datamissing}
	\footnotesize
	\setlength{\tabcolsep}{1.8pt} 
		\begin{tabular}{ccccccccccccccccc}
			\hline
			\multirow{2}{*}{Row} & \multirow{2}{*}{Real-time} & \multicolumn{3}{c}{Data} & \multicolumn{4}{c}{Track Error (km)} & \multicolumn{4}{c}{Pressure Error (hPa)} & \multicolumn{4}{c}{Wind Speed Error (m/s)} \\
			& & $X_{his}$ & $X_{ERA5}$ & $X_{tilt}$ & 
			+6h & +12h & +18h & +24h &
			+6h & +12h & +18h & +24h &
			+6h & +12h & +18h & +24h \\
			\hline
			1 & yes & / & / & / & 114.88 & 228.64 & 331.53 & 428.40 & 2.78 & 4.93 & 6.04 & 6.94 & 1.57 & 2.84 & 3.72 & 4.50 \\
			2 & yes & 6h & / & / & 57.30 & 104.16 & 146.70 & 191.52 & 2.20 & 3.08 & 3.62 & 4.15 & 1.34 & 1.93 & 2.26 & 2.58 \\
			3 & yes & 12h & / & / & 34.71 & 57.52 & 83.13 & 116.12 & 1.31 & 1.94 & 1.91 & 2.62 & 0.62 & 1.25 & 1.17 & 1.54 \\
			4 & yes & 18h & / & / & 26.43 & 39.97 & 61.88 & 93.99 & 1.10 & 2.03 & 1.97 & 2.67 & 0.59 & 1.13 & 1.09 & 1.47 \\
			5 & yes & 24h & / & / & 20.97 & 30.21 & 46.45 & 77.19 & \textbf{0.96} & 1.60 & 1.61 & 2.33 & 0.50 & 0.86 & 0.86 & 1.25 \\
			6 & yes & 30h & / & / & 19.12 & 26.48 & 43.10 & 74.86 & 0.99 & 1.61 & 1.61 & 2.28 & 0.50 & 0.83 & 0.82 & 1.23 \\
			7 & yes & 36h & / & / & 18.21 & 23.60 & 39.81 & 71.61 & \textbf{0.96} & 1.54 & \textbf{1.55} & \textbf{2.23} & \textbf{0.49} & 0.81 & \textbf{0.80} & 1.21 \\
			8 & yes & 42h & / & / & 17.99 & 22.77 & 38.70 & 70.16 & \textbf{0.96} & 1.53 & 1.56 & 2.24 & \textbf{0.49} & 0.81 & 0.81 & \textbf{1.20} \\
			9 & yes & 48h & / & / & \textbf{17.80} & \textbf{21.89} & \textbf{37.51} & \textbf{69.03} & \textbf{0.96} & \textbf{1.52} & \textbf{1.55} & \textbf{2.23} & \textbf{0.49} & \textbf{0.80} & \textbf{0.80} & \textbf{1.20} \\
			10 & no & 48h & \checkmark & / & \textbf{17.80} & \textbf{21.89} & \textbf{37.51} & \textbf{69.03} & \textbf{0.96} & \textbf{1.52} & \textbf{1.55} & \textbf{2.23} & \textbf{0.49} & \textbf{0.80} & \textbf{0.80} & \textbf{1.20} \\
			11 & no & 48h & / & \checkmark & \textbf{17.80} & 21.91 & 37.55 & 69.08 & \textbf{0.96} & \textbf{1.52} & \textbf{1.55} & \textbf{2.23} & \textbf{0.49} & \textbf{0.80} & \textbf{0.80} & \textbf{1.20} \\
			12 & no & 48h & \checkmark & \checkmark & \textbf{17.80} & 21.91 & 37.55 & 69.08 & \textbf{0.96} & \textbf{1.52} & \textbf{1.55} & \textbf{2.23} & \textbf{0.49} & \textbf{0.80} & \textbf{0.80} & \textbf{1.20} \\
			\hline%
		\end{tabular}
\end{table}
\begin{figure}[!htb]
	\centering 
	\begin{tabular}{c}
		\includegraphics[width=0.78\textwidth]{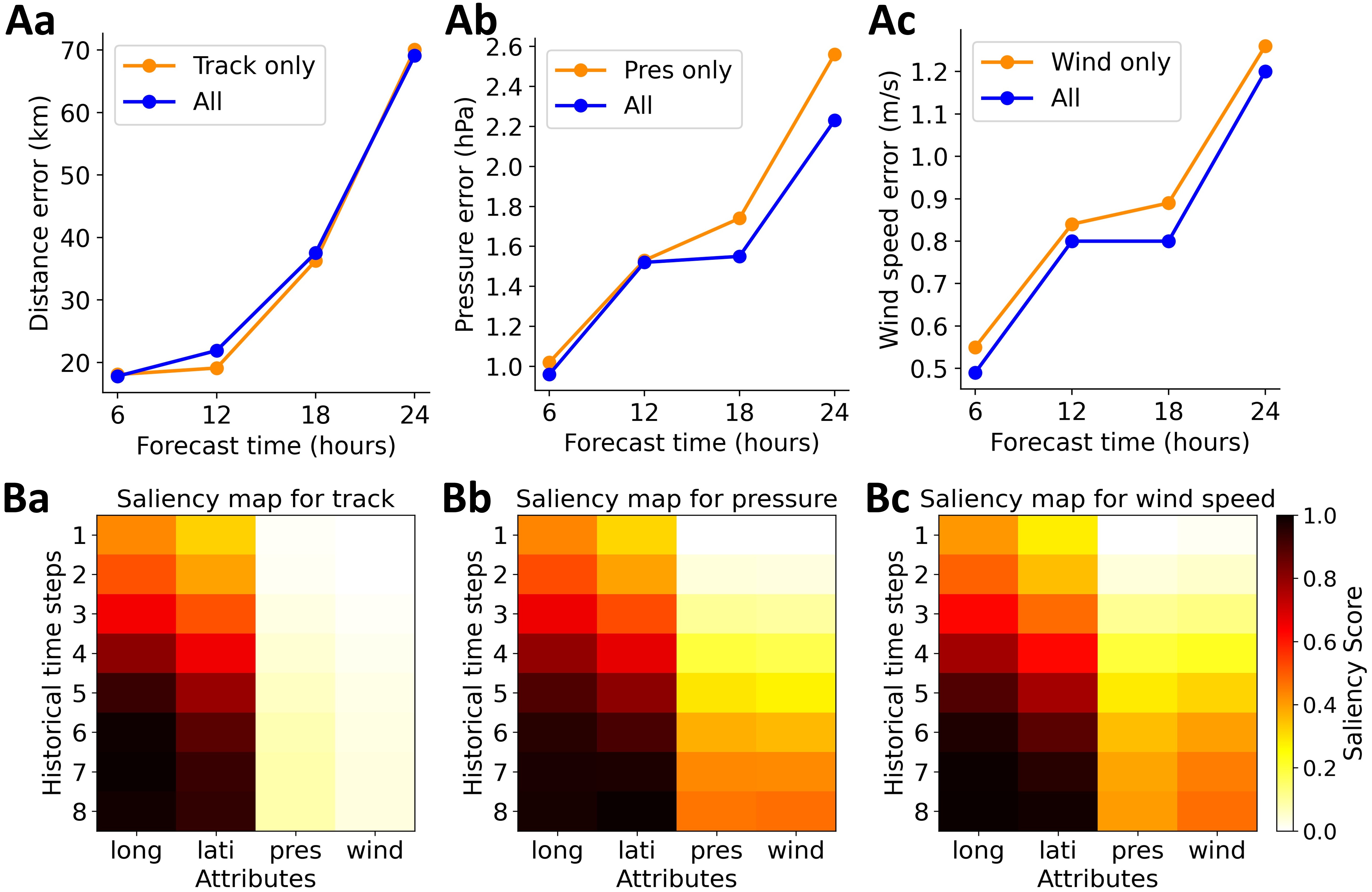}\\
	\end{tabular}	
	\caption{\textbf{Analyzing the correlation between track and intensity prediction.} (A) Ablation results when predictions are made individually and jointly. (B) Saliency maps of historical attribute values. The columns correspond to (\textbf{a}) track, (\textbf{b}) pressure, and (\textbf{c}) wind speed.}	
	\label{fig:multitask}
\end{figure}

\subsection*{Intensity prediction is dependent on track}  

A core challenge in tropical cyclone forecasting is the simultaneous prediction of both track and intensity. Traditionally, these two tasks have been modeled independently, reflecting the fact that they are governed by different dominant physical mechanisms:   
track prediction is mainly influenced by the steering flow, such as the position and strength of the subtropical high~\cite{wu2020interannual},  
while intensity prediction depends more heavily on factors like sea surface temperature, vertical wind shear, and ocean-atmosphere interactions~\cite{wong2004tropical, schade2000tropical}. 

However, growing evidence suggests a strong coupling between track and intensity. In our proposed approach, both attributes are predicted jointly in an end-to-end framework, allowing the model to learn potential cross-task dependencies.   
As shown in Figure~\ref{fig:multitask}A, this joint prediction leads to superior performance compared to separate forecasting.  
Furthermore, the saliency analysis in Figure~\ref{fig:multitask}B highlights how intensity forecasts are sensitive to track-related features, providing further evidence for the interdependence between these two TC attributes.

A key conclusion is that while TC track and intensity are physically interconnected, their dependencies are asymmetric. 
Track prediction appears to be largely independent of intensity, whereas intensity prediction is strongly dependent on the track.  
As illustrated in Figure~\ref{fig:multitask}, this is primarily because the environmental field is the dominant driver of TC intensity changes.  
As a TC moves along its track, it encounters different environmental conditions (e.g., ocean heat content, wind shear), which in turn influence its structure and intensity.  
Thus, accurate track forecasts provide crucial context for predicting intensity evolution, reinforcing the necessity of jointly modeling both attributes.

\section*{Discussion}\label{sec12}
As shown in Table~\ref{tab:72hours_error}, Tianmu-TC's intensity forecasting maintains a clear advantage over the China Meteorological Observatory (CMO), and the European Centre for Medium-Range Weather Forecasts global model, even at the +72 h forecast horizon. However, it must be acknowledged that this work still has some limitations. Track forecasting performance becomes suboptimal beyond 24 h, mainly due to the lack of explicit atmospheric dynamic constraints, and limited long-range temporal dependency modeling capability. Addressing these issues will be the focus of our future work.

\begin{table}[h]
	\caption{\textbf{Compare the prediction errors at a 72-hour forecast horizon with authoritative NWP methods in WP basin.} The testing years are from 2017 to 2023. 
	}\label{tab:72hours_error}
	\begin{tabular*}{\textwidth}{@{\extracolsep\fill}l lccccc}
			\hline
			& & \multicolumn{5}{c}{Prediction Horizon (hours)} \\\cline{3-7}
			Error Type & Method & $+6h$ & $+12h$ & $+24h$ & $+48h$ & $+72h$ \\
			\hline
			\multirow{3}{*}{\textbf{Track Error (km)}}  
			& CMO & 32.4 & 52.9 & 75.5 & 132.1 & 201.1 \\ 
			& ECMWF & 42.3 &	47.0 &	\textbf{65.8} &	\textbf{116.8} &	\textbf{193.6}  \\
			&\textbf{Tianmu-TC} & \textbf{17.8} & \textbf{21.9} & 69.1 & 238.7 & 438.9 \\
			\hline
			\multirow{3}{*}{\textbf{Pressure Error (hPa)}}  
			& CMO &2.6&	4.37	&6.25&	9.01&	11.25  \\ 
			& ECMWF   &10.9 &	11.2 &	11.8 &	12.9 &	11.7 
			 \\
			&\textbf{Tianmu-TC} & \textbf{1.0} &\textbf{1.5} &\textbf{2.1} &\textbf{4.8} &\textbf{6.2} \\
			\hline
			\multirow{3}{*}{\textbf{Wind Speed Error (m/s)}}  
			& CMO & 1.5&	2.3	&3.5&	4.9&	6.0 \\  
			& ECMWF
			& 7.2 	&7.2 &	7.2 &	7.4 &	6.9 \\   
			& \textbf{Tianmu-TC} & \textbf{0.5} & \textbf{0.8} & \textbf{1.2} & \textbf{3.0} & \textbf{4.0} \\  
			\hline
		\end{tabular*}
\end{table}

\section*{Materials and Methods}
\subsection*{Experimental Setup}
\label{sec:Experimental Setup}
For fair comparison, following TCN$_{M}$~\cite{huang2025benchmark}, the sampling number is 6, which means that the model generates 6 possible tendencies. Similarly, Tianmu-TC inputs 48 hours ($n$ = 8, the time resolution is 6 hours) of TC historical data and outputs 24 hours ($m$ = 4) of future TC attribute data. Tianmu-TC was deployed on the PyTorch framework. Training was performed using Adam Optimizer with a learning rate of 0.0001, a batch size of 256, and a duration of 48 hours. 

All experiments, including other deep learning methods used for comparison, were conducted on an NVIDIA RTX A6000 GPU. Random seeds were fixed during both training and testing to ensure reproducibility. The number of training epochs was set to 350 based on model convergence criteria. 
Absolute errors were computed between predicted results and ground truth for track (distance in km), pressure (hPa), and wind speed (m/s).

\subsection*{Data preparation and availability} 
We organize a new dataset, which mainly consists of three parts: historical TC attributes, ERA5 data representing external environment, and TC tilt metrics representing TC internal structure. This dataset encompasses \underline{all} the 1786 TCs data from 1950 to 2023 over the Western North Pacific. So the datasets contain sufficiently diverse conditions of TCs. 80\% of the TC data from 1950 to 2016 was allocated for training, 20\% for validation, and the data from 2017 to 2023 were reserved for testing. Environmental data from MGTCF—including future move direction, intensity class and future intensity change direction—were also utilized. Further details are described below.

\textbf{Historical TC attributes} are derived from International Best Track Archive for Climate Stewardship (IBTrACS~\cite{Knapp2010ibtracs}), including essential TC center information such as longitude ($^\circ$), latitude ($^\circ$), pressure (hPa) and wind speed (m/s). Time resolution is 6 hours.

\textbf{ERA5 data} are derived from the fifth-generation atmospheric reanalysis of the global climate (ERA5)~\cite{hersbach2018era5} by the European Centre for Medium-Range Weather Forecasts (ECMWF). We selected the geopotential height data at 200 hPa and 500 hPa, represented as $z200$ and $z500$, which indicate high-pressure systems, as well as the large-scale wind field data, including $u200$, $u500$, $u700$, $u850$, $v200$, $v500$, $v700$, and $v850$. Here, $u$ represents the zonal wind component (east-west direction), and $v$ represents the meridional wind component (north-south direction). The 200 hPa data reflect upper-level information of tropical cyclones, the 500 hPa data represent mid-level information, while the 700 hPa and 850 hPa data capture lower-level (near-surface) information of tropical cyclones. The image size is 20$^\circ$ $\times$ 20$^\circ$, centered on the TC eye, covering a radius of 10$^\circ$ around the TC's eye. The spatial resolution is 0.25$^\circ$, resulting in a data size of 81 $\times$ 81.	

\textbf{TC tilt metrics} are derived from vorticity data, which are calculated from wind field data. As shown in Figure~\ref{fig:3constraints}D, vorticity at 200 hPa, 500 hPa, and 850 hPa are first computed using the corresponding $u$ and $v$ components for each pressure level, as described in Equation~\ref{equ:vo}. The resulting vorticity data for these three pressure levels are referred to as $vo200$, $vo500$, and $vo850$, respectively. Next, taking $vo200$ as an example, as shown in Figure~\ref{fig:3constraints}D, the image size is 81 $\times$ 81. The TC eye is marked by a black diamond at the center of the image, located at position (41, 41), denoted as ($\diamondsuit_x$, $\diamondsuit_y$). The three local minima closest to the TC eye are marked as red circles, and their average coordinates are calculated to obtain the center of the three local minima, which is marked as a blue triangle ($\bigtriangleup_x^{pl}$, $\bigtriangleup_y^{pl}$). $pl\in\{200,500,850\}$ refers to pressure level.
\begin{equation}
	\label{equ:vo}
	vo = \frac{\partial v}{\partial x} - \frac{\partial u}{\partial y}
\end{equation}
\begin{equation}
	\label{equ:dx}
	\partial x = \partial y \times \cos\left(\frac{\pi}{180} \times \text{avg\_latitude}\right)
\end{equation} 
where $\partial y = 111.32 \times 0.25$, 111.32 represents the distance in kilometers corresponding to 1 degree of longitude. The value 0.25 refers to the spatial resolution. Thus, \(\partial y\) represents the distance between data points in the longitudinal direction, with units in kilometers. Similarly, \(\partial x\) represents the distance between data points in the latitudinal direction, which varies depending on the latitude. Here, it is simplified to the average latitude (\(\text{avg\_latitude}\)). Since the training data is based on tropical cyclones in the WP, the average latitude is set to 15$^\circ$.
\begin{equation}
	\label{equ:dcen200}
	D_{\text{cen200}} = \sqrt{\left(\bigtriangleup_x^{200} - \diamondsuit_x\right)^2 + \left(\bigtriangleup_y^{200} - \diamondsuit_y\right)^2}
\end{equation}
\begin{equation}
	\label{equ:dcen850}
	D_{\text{cen850}} = \sqrt{\left(\bigtriangleup_x^{850} - \diamondsuit_x\right)^2 + \left(\bigtriangleup_y^{850} - \diamondsuit_y\right)^2}
\end{equation}
\begin{equation}
	\label{equ:d28}
	D_{\text{28}} = \frac{1}{1 + \sqrt{\left(\bigtriangleup_x^{200} - \bigtriangleup_x^{850}\right)^2 + \left(\bigtriangleup_y^{200} - \bigtriangleup_y^{850}\right)^2}}
\end{equation}
\begin{equation}
	\label{equ:d58}
	D_{\text{58}} = \frac{1}{1 + \sqrt{\left(\bigtriangleup_x^{500} - \bigtriangleup_x^{850}\right)^2 + \left(\bigtriangleup_y^{500} - \bigtriangleup_y^{850}\right)^2}}
\end{equation}
where $D_{28}$ represents the alignment between 200 hPa and 850 hPa; $D_{58}$ represents the alignment between 500 hPa and 850 hPa. $D_{\text{cen200}}$ and $D_{\text{cen850}}$ measure the distance between the TC eye and local minima center at 200 hPa and 850 hPa, respectively. The same calculations were performed for each historical time step, resulting in a sequential series of TC tilt metrics.

\subsection*{Tianmu-TC network design}
Overall, Tianmu-TC can be divided into three main modules: constraint encoder, cross-modal fusion, and constraint guiding denoising. The constraint guiding denoising consists of $K$ Physics-constraints Denoising Blocks.

\textbf{Constraint Encoder: How to design the 3 constraints?}

\textbf{LSTM-based Encoder for encoding TC track and intensity}, which is designed based on the encoder from Trajectron++~\cite{salzmann2020trajectron++}. The input data has a shape of $(T, 4)$, where $T = 8$, representing the historical attributes of the tropical cyclone over $8 \times 6$ hours $= 48$ hours. The four values correspond to longitude, latitude, pressure, and wind speed. After encoding, the output $F_1$ has a shape of $(256)$.

\textbf{CNN-based Encoder} encodes selected ERA5 data. The data includes $u200$, $u500$, $u700$, $u850$, $v200$, $v500$, $v700$, $v850$, $z200$, and \( z500\), totaling 10 variables. Therefore, the data shape is \((T, 10, 81, 81)\), where \(T = 8\). The predicted attributes are all the attributes in and around the TC center, which we define as a highly task-relevant area. To enhance task-relevant information, the Central Information Enhancement (CenIE) block enhances the center point of the original ERA5 data (see in the Figure~\ref{fig:KIE+2D-enc} of supplementary materials). As a result, the input data size becomes \((T, 20, 81, 81)\). We implement channel fusion through 6 layers of CNN with ReLU activation functions and 3 layers of spatial attention. This process outputs data with a shape of \((T, 1, 16, 16)\), where the channel dimension \(C\) is reduced from 20 to 1, and the height (\(H\)) and width (\(W\)) are downscaled.

\textbf{Another LSTM-based encoder encodes TC tilt metrics}. The input data has a shape of \((T, 4)\), where \(T = 8\). The 4 values represent four metrics: \(D_{\text{cen200}}, D_{\text{cen850}}, D_{28}, D_{58}\). A single-layer LSTM is used to encode the data, resulting in \(F_3\) with a size of \(256\).

\textbf{Cross-modal Fusion: Establishing relationships among the three constraints.}  

The outputs from the constraint encoder have shapes of $F_1$ (256), $(T, 1, 16, 16)$, and $F_3$ (256), respectively. From the input data shapes, it can also be observed that TC track and intensity, as well as TC tilt values, are one-dimensional attribute values, while ERA5 data is two-dimensional image data. Therefore, it is necessary to match and fuse these two modalities. 

As shown in Figure~\ref{fig:framework}, we select the encoding result (256) of TC track and intensity, which is semantically most relevant to the prediction target, and first reshape it into a shape of $(1, 16, 16)$. We initialize $H$ to 0, with a shape of $(1, 16, 16)$. 

The iteration is performed over each historical time step, with $T$ ranging from 1 to 8. At $T=1$, the corresponding slice from $(T, 1, 16, 16)$ with shape $(1, 16, 16)$ is selected and concatenated with the reshaped encodings of TC track, intensity, and $H$, yielding a final shape of $(3, 16, 16)$. This is then sent into the \textbf{\textit{SA}} (self-attention module) to capture semantic associations among the three, producing a new $H$ with a shape of $(1, 16, 16)$. This process is repeated to update $H$ until $T=8$. After reshape the final $H$, the feature size is (256), which is the result of temporal fusion of ERA5 data, denoted as $F_2$. 

Finally, a 4-layer Transformer encoder is used to fuse $F_1$, $F_2$, and $F_3$ to obtain $C_1$, $C_2$, and $C_3$. In fact, these three constraints are fused into a complete feature of size (256), denoted as $C$. 

\textbf{Constraint guiding denoising}

A diffusion sequence is $(y_0, y_1, ..., y_K)$, where $K$ is the maximum diffusion step. The diffusion process gradually adds noise into the ground truth region $y_0$ of TC forecasting. Correspondingly, the reverse diffusion sequence is $(y_K, y_{K-1}, ..., y_0)$. This process utilizes historical data as conditions, and gradually denoises the standard Gaussian $y_K$ to obtain the 1D future TC attributes $y_0$. This process reduces prediction noise.

\textbf{The diffusion process} is defined as $q(y_{k}|y_{0}):=\mathcal N(y_k; \sqrt{\bar{\alpha_k}}y_{0}, (1-\bar{\alpha_k})\mathrm I)$, thus
\begin{equation}
	y_{k}=\sqrt{\bar{\alpha_k}}y_0+\sqrt{(1-\bar{\alpha_k})}\varepsilon
\end{equation}
$\bar{\alpha_k}=\prod_{s=1}^{k}\alpha_s$, $\alpha_1,\alpha_2,...\alpha_k$ are fixed variance schedulers, the noise variable $\varepsilon\sim\mathcal N(0,\mathrm I)$. As the diffusion step $k$ increases, $y_k\sim\mathcal N(0,\mathrm I)$. This means $z_0$ is gradually destroyed into a Gaussian noise distribution. The training loss is:
\begin{equation}
	\setlength{\abovedisplayskip}{2pt}
	\setlength{\belowdisplayskip}{3pt}
	L(\theta,\psi)=\mathbb{E}_{\varepsilon,z_0,k}||\varepsilon-\varepsilon_{(\theta,\psi)}(y_k,k,X)||
\end{equation}
$\theta$ are parameters of Physics-constraints Denoising model, $\psi$ are parameters of constraint encoder and cross-modal fusion, $X$ are the input of the model.

The generation of increasingly reliable TC predictions is formulated as a \textbf{reverse diffusion process}, guided by three specifically designed constraints that progressively reduce uncertainty.

The reverse diffusion process with Bi-condition is defined as:
\begin{equation}
	\textstyle p_{\theta}(y_{k-1}|y_k,C):=\mathcal N(y_{k-1};\mu_{\theta}(y_k,k,C);\Sigma_{\theta}(y_k,k))
\end{equation}
\begin{equation}
	\mu_{\theta}(\cdot)=\frac 1 {\sqrt{\alpha_k}}(y_k-\frac {\beta_k} {\sqrt{1-\bar{\alpha_k}}} \varepsilon_{\theta}(y_k,k,C))
\end{equation}
\begin{equation}
	\Sigma_{\theta}(\cdot)=\sigma_k^2\mathrm I=\beta_k\mathrm I
\end{equation}
thus:
\begin{equation}
	\label{equ:yk_1}
	y_{k-1}=\frac{1}{\sqrt{\alpha_k}}\left(y_k-\frac{\beta_k}{\sqrt{1-\bar{\alpha_k}}}\varepsilon_{\theta}(y_k,k,C)\right)+\sqrt{\beta_k}e
\end{equation}
In Equation~\ref{equ:yk_1}, $e$ is a random variable in standard Gaussian Distribution, $\beta_k=1-\alpha_k$. $\varepsilon_{\theta}$ is implemented by the Physics-constraints denoising block in Figure~\ref{fig:framework}. The output of the block is:
\begin{equation}
	y_{k}'=\varepsilon_{\theta}(y_k,k,C)
\end{equation}
inspired by MID~\cite{gu2022stochastic}, $\varepsilon_{\theta}$ is implemented mainly by a Transformer encoder and 4-layer gating units, which use constraints $C$ to generate Gate and bias, and perform a gating transform for latent noise $y_k$.





\clearpage 

%


\section*{Acknowledgments}
\paragraph*{Funding:}
This work is partially supported by Zhejiang Provincial Natural Science Foundation of China under Grant No. RG25F020001 and LR21F020002, as well as the Natural Science Foundation of China under Grant No. 62202429.

\paragraph*{Author contributions:}
Conceptualization: S.Z., P.M., C.H., C.B., and S.S. Methodology: S.Z., P.M., S.S., C.H., H.Y., and Y.Z. Software: S.Z., P.M., Y.Z., C.H. Validation: S.Z., P.M., H.Y., Y.Z., C.B. and S.S. Formal analysis: S.Z., P.M., C.B., S.S., H.Y., C.H., J.Zh., and S.C. Investigation: S.Z., P.M., C.H., S.S., and C.B. Resources: C.B., S.S., S.C., and J.Zh. Data curation: S.Z., C.H., Y.Z., and H.Y. Writing—original draft: S.Z., P.M. Writing—review and editing: C.B., S.S., P.M., C.H., J.Zh., and S.C. Visualization: S.Z., P.M., and S.S. Supervision: C.B. and S.S. Project administration: C.B. and S.S. Funding acquisition: C.B.

\paragraph*{Competing interests:}
There are no competing interests to declare.
\paragraph*{Data and materials availability:}
Our dataset consists of three main components:

\begin{enumerate}
	\item \textbf{Historical TC attributes}, sourced from our previously organized dataset, available at https://zenodo.org/records/15009527.
	\item \textbf{ERA5 data}, provided by ECMWF, which can be accessed from the https://cds.climate.copernicus.eu/.
	\item \textbf{TC tilt metrics}, derived from vorticity data, which are computed using wind field data. The processing code is available at https://github.com/Zjut-MultimediaPlus/Tianmu-TC.
\end{enumerate}

In our model, these datasets are preprocessed into \textbf{PKL} files, which can be obtained from Google Drive: https://drive.google.com/file/d/1XpfByEZkZHAybXgB5p2YsR5KZhHrtVei/view?usp=drive\_link and https://drive.google.com/file/d/1aiJaUH035YOIbsS9Q1Y9GGmyKW1HiJU1/view?usp=drive\_link.

All the code and processed data are public on Github~( https://github.com/Zjut-MultimediaPlus/Tianmu-TC). This code will be updated and changed over time.


\subsection*{Supplementary materials}
Supplementary Text\\
Table S1\\
Figures S1 to S4\\
Movies S1 to S3\\


\newpage


\renewcommand{\thefigure}{S\arabic{figure}}
\renewcommand{\thetable}{S\arabic{table}}
\renewcommand{\theequation}{S\arabic{equation}}
\renewcommand{\thepage}{S\arabic{page}}
\setcounter{figure}{0}
\setcounter{table}{0}
\setcounter{equation}{0}
\setcounter{page}{1} 

\appendix
\begin{center}
\section*{Supplementary Materials for\\ \scititle}
Shiqi~Zhang$^{\dagger}$,
Pan~Mu$^{\dagger}$,
Cheng~Huang,
Hanting~Yan,
Yuchao~Zhu,\\
Jinglin~Zhang,
Shengyong~Chen$^{\ast}$,
Shoujuan~Shu$^{\ast}$,
Cong~Bai$^{\ast}$\\ 
\small$^\ast$Corresponding author. Email: csy@tjut.edu.cn, sjshu@zju.edu.cn, congbai@zjut.edu.cn\\
\small$^\dagger$These authors contributed equally to this work.
\end{center}

\subsubsection*{This PDF file includes:}
Supplementary Text\\
Table S1\\
Figures S1 to S4\\
Captions for Movies S1 to S3

\subsubsection*{Other Supplementary Materials for this manuscript:}
Movies S1 to S3\\
\newpage


\subsection*{Supplementary Text}
\setcounter{secnumdepth}{3}

\renewcommand{\thesection}{\Alph{section}}
\renewcommand{\thesubsubsection}{\thesection.\arabic{subsubsection}}

\setcounter{section}{1}
\setcounter{subsubsection}{0}

\subsubsection{Diffusion Models for Multi-Trend Generation}
Diffusion models are inherently well-suited for multi-trend generation due to their ability to capture complex, multi-modal data distributions. Their probabilistic framework allows for flexible sampling, enabling the generation of diverse trends by conditioning on specific features. Additionally, the iterative refinement mechanism in the reverse diffusion process ensures scalability and accurate representation of overlapping patterns. Empirical successes in generative tasks further validate their compatibility with multi-trend dynamics, making diffusion models a natural choice for such challenges.

For fair comparison, following TCN$_{M}$~\cite{huang2025benchmark}, the sampling number is 6, which means that the model generates 6 possible tendencies. Similarly, Tianmu-TC inputs 48 hours ($n$ = 8, the time resolution is 6 hours) of TC historical data and outputs 24 hours ($m$ = 4) of 1D future TC attribute data. 

\subsubsection{NeuralGCM Tropical Cyclone Tracking} 
\label{supple:neuralgcm} 

NeuralGCM provides publicly available forecast data for 2020 (\url{https://neuralgcm.readthedocs.io/en/latest/neuralgcm_datasets.html}). In this study, tropical cyclone tracking is performed according to the available forecast variables and the tracking parameters reported in Table I7 of the original NeuralGCM paper.
The original NeuralGCM setting uses sea level pressure rather than MSLP, while the released forecast variables used in this study do not include SLP. Therefore, NeuralGCM cannot be used for intensity prediction and is only evaluated for track prediction. 

For the deterministic forecast, we use the finest-resolution 0.7° NeuralGCM forecast data provided officially (\url{gs://weatherbench2/datasets/neuralgcm_deterministic}). According to Table I7 of the original NeuralGCM paper, vorticity tracking is used for tropical cyclone tracking. The criteria are: vorticity at 850 hPa change of \(\pm 0.00006~\mathrm{s}^{-1}\) over 5.5 GCD, together with a difference between geopotential surfaces at 300 and 500 hPa that decreases by at least \(-25.8~\mathrm{m}^{2}\mathrm{s}^{-2}\) over 6.5 GCD. Here, GCD denotes great-circle distance. For the ensemble forecast (\url{gs://weatherbench2/datasets/neuralgcm_ens}), due to the high computational cost, the original NeuralGCM paper reduced the highest resolution to 1.4° for ensemble forecasts. The tracking method is consistent with that used for the deterministic forecast, and the ensemble size is 50 members. 

\subsubsection{GenCast Tropical Cyclone Tracking and Intensity Extraction}
\label{supple:gencast}
This section describes how tropical cyclone centers and intensity variables were extracted from the public GenCast forecast archive (\url{gs://weatherbench2/datasets/gencast}). We first summarize the original TempestExtremes tracker used in GenCast, then describe the modifications required by the limited pressure-level variables available in WeatherBench2, followed by the candidate review procedure and the final intensity extraction strategy.

The original GenCast used TempestExtremes v2.1 to track tropical cyclones. The tracker consists of two main stages. In the first stage, DetectNodes identifies candidate cyclone centers at each time step. Candidate locations are first determined from local minima in mean sea level pressure (MSL), and each low-pressure center is required to be associated with a sufficiently compact closed low-pressure structure. In addition, the original tracker requires the candidate storm to be colocated with an upper-level warm-core structure in the 500--300 hPa geopotential thickness field. In the second stage, StitchNodes links candidate nodes across time into physically plausible cyclone trajectories, and further filters non-tropical-cyclone trajectories using criteria related to the maximum displacement between adjacent time steps, the maximum temporal gap, the minimum track duration, wind speed, surface elevation, and latitude. Because GenCast forecasts are evaluated at 12 h intervals, the original study adjusted the StitchNodes connection range from the default 8° to 12° in great-circle distance. This larger range allows candidate centers at adjacent forecast times to be linked even when a tropical cyclone travels farther within a 12 h interval.

In this study, the public WeatherBench2 GenCast archive provides pressure-level variables only at 500, 700, and 850 hPa. The 300 hPa geopotential height required to reproduce the original 300--500 hPa warm-core criterion is therefore unavailable. As a result, the original TempestExtremes tracking protocol used in GenCast cannot be fully reproduced. To remain as consistent as possible with the original tracking rationale under the constraints of the public archive, we retained the core DetectNodes criteria based on local MSLP minima and closed low-pressure structures. Following the 12° displacement scale used for 12 h tracking in GenCast, we implemented a 12° local search window for each lead time. The search is anchored at the best-track center for the first lead time and, in the stepwise procedure, at the previously predicted center for the subsequent lead time.
It should be emphasized that the goal of this study is not to redetect the genesis, termination, or complete life cycle of all global cyclones. Instead, for each given best-track sample, we diagnose the +12 h and +24 h tropical cyclone centers and intensities from the original GenCast gridded forecast fields. Therefore, the full global StitchNodes trajectory construction, minimum-duration filtering, and elevation and latitude filtering are not applied. 

Specifically, for each initialization time and forecast lead time, MSLP, 850 hPa horizontal winds, and 10 m winds are first extracted within a 12° latitude--longitude window around the tracking anchor. The tracking anchor is initialized as the best-track center at the forecast initialization time. In the stepwise tracking procedure, the search anchor for the later lead time is updated to the predicted center from the previous lead time. Candidate centers are generated from local minima in the MSLP field, sorted by their MSLP values in ascending order, and at most 500 candidates are examined. Each candidate is then tested for the presence of a sufficiently compact closed low-pressure structure.

Because the 300 hPa warm-core criterion cannot be reproduced, 850 hPa relative vorticity is further used as an auxiliary constraint on the low-level cyclonic structure. The 850 hPa relative vorticity is computed from the 850 hPa horizontal wind field. For candidates in the Northern Hemisphere, cyclonic vorticity corresponds to a positive vorticity core, so the maximum relative vorticity within a 1° radius of the candidate center is used. For candidates in the Southern Hemisphere, cyclonic vorticity corresponds to a negative vorticity core, so the minimum relative vorticity within a 1° radius of the candidate center is used. This criterion does not require the vorticity center and the MSLP minimum to fall on exactly the same grid point; rather, it requires a consistent low-level cyclonic structure near the candidate low-pressure center.

The final center is selected using a hierarchical priority strategy. Among the four lowest local MSLP minima, candidates that satisfy both the SLP closed-contour criterion and the 850 hPa vorticity-structure criterion, and that are close to the tracking anchor, are given the highest priority. All candidates are retained, and whether each candidate passes the SLP and SLP+VORT criteria is recorded for subsequent quality diagnosis. 

Because this tracker cannot fully reproduce the TempestExtremes stitching procedure used in the original GenCast, we further performed candidate-level quality review of the automatic tracking results. For every 2020 test sample and lead time, we saved the candidate table, MSLP field, 850 hPa vorticity field, 10 m wind field, and the corresponding visualizations. The review was based on three criteria. First, the selected center should be a local MSLP minimum located within the innermost closed low-pressure structure, rather than a grid point on the outer edge of the closed low or along a low-pressure trough. Second, the candidate should be associated with a consistent cyclonic vorticity core in the 850 hPa relative vorticity field, namely a local positive-vorticity maximum in the Northern Hemisphere and a local negative-vorticity minimum in the Southern Hemisphere. Third, the predicted centers at +12 h and +24 h should be temporally continuous, avoiding jumps to unrelated low-pressure systems between adjacent lead times. For the small number of samples with clear jumps in the automatic tracking results, corrections were restricted to selecting, from the candidate table saved during automatic tracking, a candidate center satisfying the above physical-structure and temporal-continuity criteria. No new center coordinates were manually specified outside the saved candidate set. To ensure traceability, we saved and provided the predicted tracks, track errors, candidate tables, and visualizations of MSLP, 850 hPa relative vorticity, and 10 m wind speed for all 2020 test samples. Overall, the reviewed centers are consistent with the closed MSLP structure, the low-level cyclonic vorticity structure, and the temporal continuity between adjacent lead times. 

After the diagnostic center is determined, GenCast intensity diagnostics are further extracted. The intensity task in this study includes two variables: pressure and wind speed. Pressure refers to the lowest atmospheric pressure at the TC center, in hPa, and wind speed refers to the two-minute maximum sustained wind speed near the TC center, in m/s. For the GenCast gridded forecast fields, we extract the minimum MSLP within a 1° radius of the diagnostic center as the minimum central pressure diagnostic, and the maximum 10 m wind speed within a 3° radius as the near-center wind speed diagnostic. It should be noted that the public GenCast archive provides instantaneous gridded wind fields at discrete forecast times, rather than two-minute maximum sustained wind observations. Therefore, the GenCast wind speed error reported in this study should be interpreted as the diagnostic error of the near-center maximum wind speed derived from the instantaneous 10 m wind field, rather than a strict error of the two-minute maximum sustained wind speed.

\begin{table}[htb]
	\centering
	\caption{\textbf{Extended-Sample Comparison on special TC scenarios on global area from 2017 to 2021.} 
		Samples is the number of cases. Lower values indicate better performance. ``Anomaly'' consists of circular movement, turning on the ocean, and V-shaped tracks. ``RI'' denotes \textit{Rapid Intensification}, and ``RW'' denotes \textit{Rapid Weakening}.}
	\label{tab:special_scenarios_extend}
	\footnotesize
	\setlength{\tabcolsep}{2.6pt}
	\begin{tabular}{lllcccccc}
		\hline
		\multirow{2}{*}{Category} & \multirow{2}{*}{Scenario (Unit)} & \multirow{2}{*}{Methods} & \multirow{2}{*}{Samples} & \multicolumn{4}{c}{Forecast Horizon} \\
		\cline{5-8}
		&  &  &  & +6h & +12h & +18h & +24h \\
		\hline
		\multirow{6}{*}{\textbf{Track}} 
		& \multirow{3}{*}{Anomaly (km)} 
		& Pangu  & 1103 & 56.25 & 62.96 & 77.58 & 110.94 \\
		& & Fengwu & 1103 & 44.19 & 49.33 & 65.89 & 95.62 \\

		& & Ours   & 1103 & \textbf{18.08} & \textbf{17.72} & \textbf{36.93} & \textbf{73.57} \\
		\cline{2-8}
		& \multirow{3}{*}{Landfall (km)} 
		& Pangu  & 582 & 59.84 & 76.14 & 101.26 & 131.16 \\
		& & Fengwu & 582 & 52.33 & 62.04 & 76.14 & 93.80 \\
		& & Ours   & 582 & \textbf{17.59} & \textbf{19.06} & \textbf{38.86} & \textbf{71.78} \\
		\hline
		\multirow{6}{*}{\textbf{Intensity}} 
		& \multirow{3}{*}{RI (m/s)} 
		& Pangu  & 330 & 20.79 & 29.41 & 39.17 & 47.02 \\
		& & Fengwu & 330 & 19.92 & 27.71 & 36.82 & 43.99 \\
		& & Ours   & 330 & \textbf{1.37} & \textbf{1.31} & \textbf{4.63} & \textbf{8.04} \\
		\cline{2-8}
		& \multirow{3}{*}{RW (m/s)} 
		& Pangu  & 464 & 28.17 & 21.53 & 15.50 & 11.37 \\
		& & Fengwu & 464 & 27.42 & 20.76 & 14.48 & 10.08 \\
		& & Ours   & 464 & \textbf{2.70} & \textbf{1.70} & \textbf{3.89} & \textbf{5.78} \\
		\hline
	\end{tabular}
\end{table}

\begin{figure}[h]
	\centering 
	\setlength{\abovecaptionskip}{-0.01cm}
	\setlength{\belowcaptionskip}{-0.2cm}
	\begin{tabular}{c}
		\includegraphics[width=0.80\textwidth]{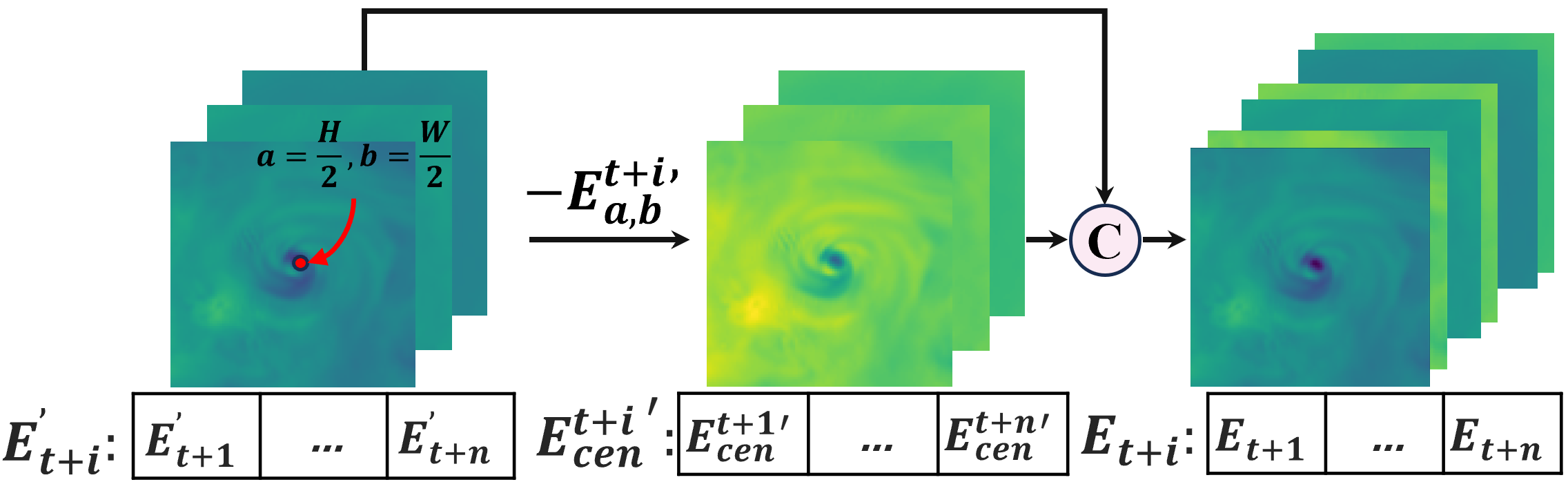}\\
	\end{tabular}
	\caption{\textbf{Central Information Enhancement (CenIE) block.} Considering that the center of selected ERA5 data corresponds to the location of the TC center, specifically, the longitude and latitude coordinates in the track task. Besides, pressure represents the lowest pressure near the TC center, and wind speed denotes the two-minute maximum sustained wind speed near the TC center. In other words, the predicted attributes are all the attributes in and around the TC center. Therefore, we define TC center as a highly task-relevant area. To enhance task-relevant information in the selected ERA5 data, CenIE block enhances the center point $E_{a,b}^{t+i'}$ of original 2D data $E_{t+i}'$ and $	E_{t+i}=concat(E_{t+i}',E_{t+i}'-E_{a,b}^{t+i'})$, where $C$ in the figure denotes channel concatenation, $i\in[1,n]$, the $a$ and $b$ denote the coordinates in the geopotential height map, $H$ and $W$ denote the width and height of the map, and $n$ denotes the length of the historical data. That is, the value of each pixel point in $E_{t+i}'$ is subtracted from that of the current central point $E_{a,b}^{t+i'}$, and concatenated with the original $E_{t+i}'$. Finally, CenIE block outputs the enhancement of central information $E_{t+i}$. This data preprocessing method is very simple and does not need GPU resources.} \label{fig:KIE+2D-enc}
\end{figure}
\begin{figure}[h]
	\centering 
	\begin{tabular}{c}
		\includegraphics[width=0.60\textwidth]{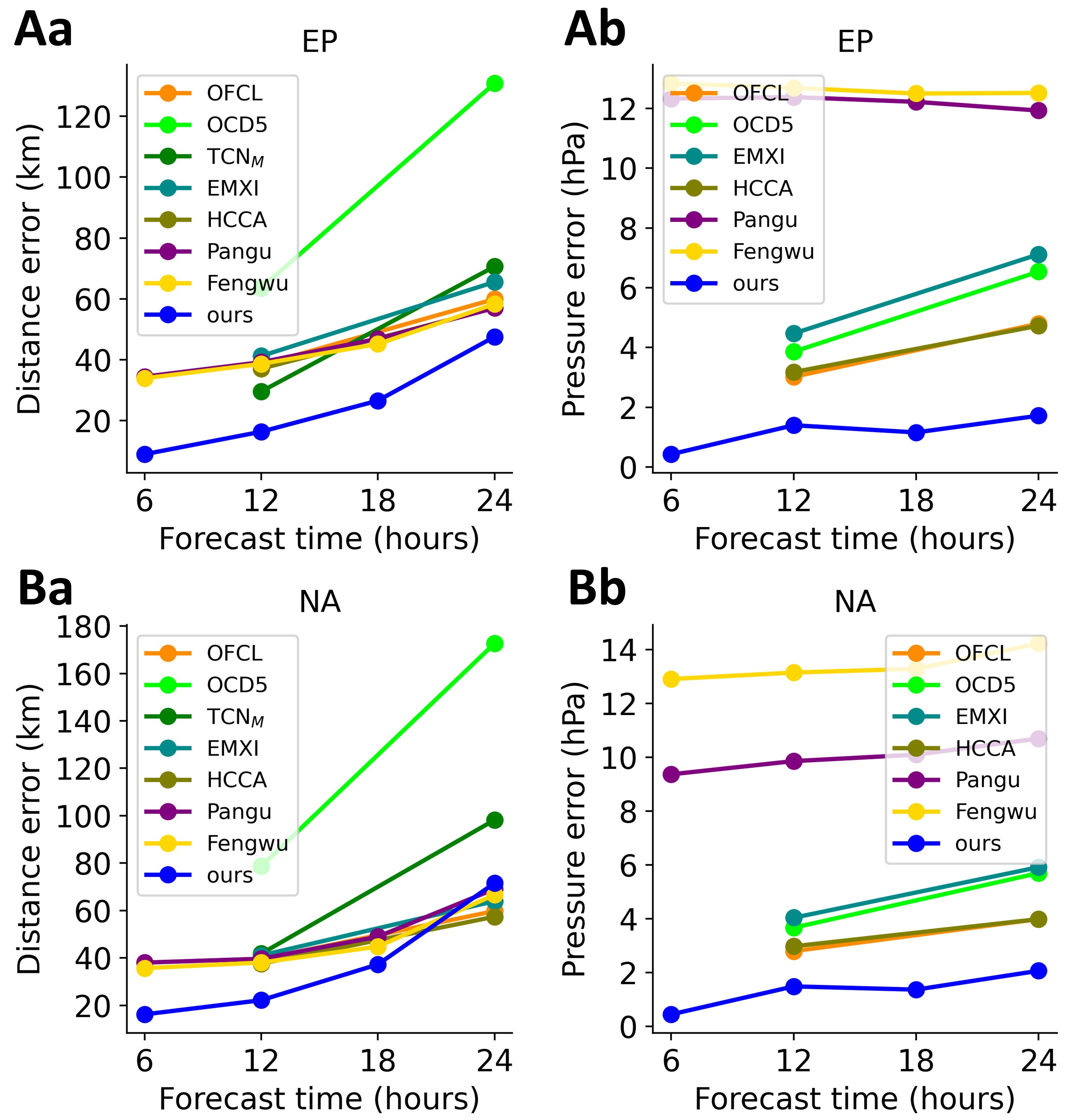}\\
	\end{tabular}
	\caption{\textbf{Comparison with additional operational models in the EP and NA basins.} 
	We include authoritative forecasting systems issued by the National Hurricane Center (NHC): the official forecast (OFCL), and the statistical-dynamical model combination OCD5, which merges CLP5 (track) and DSHP (intensity). Other models include EMXI (ECMWF global model from the previous cycle), HCCA (a weighted consensus of multiple dynamical and statistical models), as well as recent large models Pangu and Fengwu. Testing period: from 2017 to 2023.} \label{fig:supple-compare}
\end{figure}
\begin{figure}[h]
	\centering 
	\begin{tabular}{c}
		\includegraphics[width=0.96\textwidth]{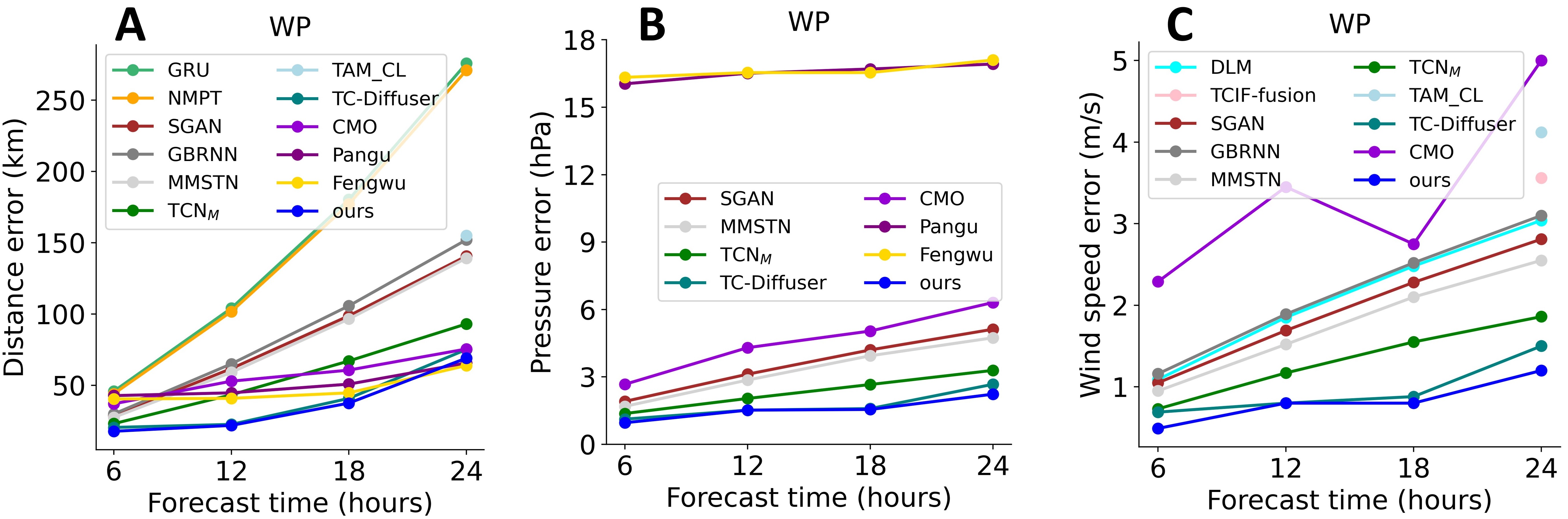}\\	
	\end{tabular}
	\caption{\textbf{More comparison methods on WP.} 
	} \label{fig:global_wp}
\end{figure}
\begin{figure}[h]
	\centering 
	\begin{tabular}{c}
		\includegraphics[width=0.96\textwidth]{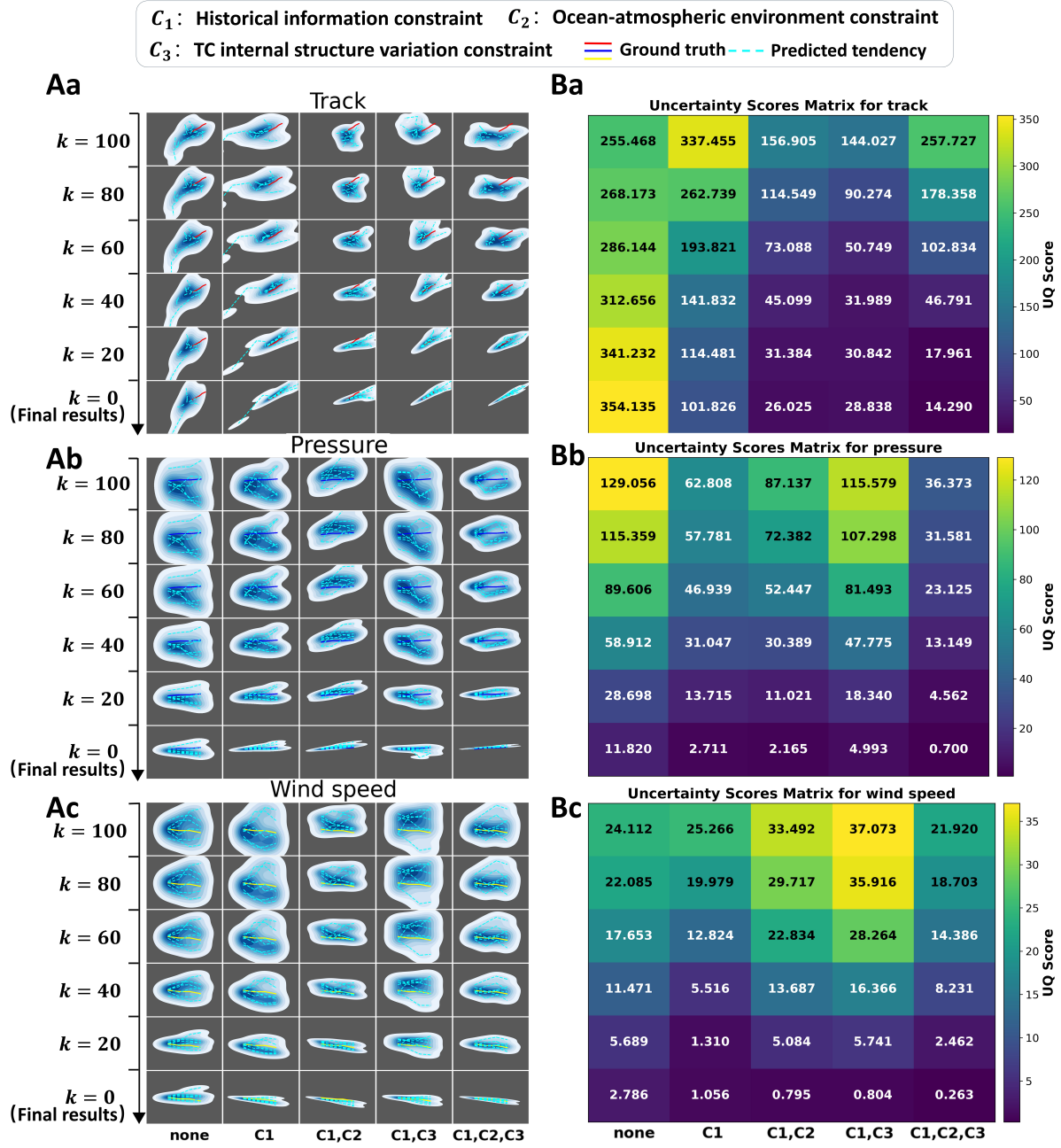}\\	
	\end{tabular}
	\caption{\textbf{Quantitative results of uncertainty for} (\textbf{a}) track, (\textbf{b}) pressure and (\textbf{c}) wind speed.} \label{fig:uq_score}
\end{figure}

\clearpage 

\paragraph{Caption for Movie S1.}
\textbf{The reduction of prediction uncertainty in track forecasting.} Animated version of Figure~\ref{fig:c1c2c3}(a). The red solid line represents the ground truth, the cyan dashed lines represent the predictions of 6 tendencies. Under the guidance of proposed physics-constrained generative AI, six tendencies gradually converge into a region with low uncertainty.

\paragraph{Caption for Movie S2.}
\textbf{The reduction of prediction uncertainty in pressure forecasting.} Animated version of Figure~\ref{fig:c1c2c3}(b). The blue solid line represents the ground truth, the cyan dashed lines represent the predictions of 6 tendencies.

\paragraph{Caption for Movie S3.}
\textbf{The reduction of prediction uncertainty in wind speed forecasting.} Animated version of Figure~\ref{fig:c1c2c3}(c). The yellow solid line represents the ground truth, the cyan dashed lines represent the predictions of 6 tendencies.

\end{document}